\documentclass[12pt]{article}

\usepackage[left=2.7cm,bottom=2.7cm,right=2.7cm,top=2.7cm]{geometry}

\usepackage{enumerate}
\usepackage{tabularx}
\usepackage{subfigure}
\usepackage{multirow}
\usepackage{algorithm}
\usepackage{authblk}
\usepackage{algorithmic}
\usepackage{indentfirst}
\usepackage{setspace}
\usepackage{color}
\usepackage[utf8]{inputenc} % allow utf-8 input
\usepackage[T1]{fontenc}    % use 8-bit T1 fonts
\usepackage{hyperref}       % hyperlinks
\usepackage{url}            % simple URL typesetting
\usepackage{booktabs}       % professional-quality tables
\usepackage{amsfonts}       % blackboard math symbols
\usepackage{nicefrac}       % compact symbols for 1/2, etc.
\usepackage{microtype}      % microtypography
\usepackage{lipsum}		% Can be removed after putting your text content
\usepackage{graphicx}
\usepackage{natbib}
\usepackage{doi}
\usepackage[namelimits]{amsmath} 
\usepackage{amssymb}            
\usepackage{amsfonts}       
\usepackage{mathrsfs} 
\usepackage[resetlabels]{multibib}
\usepackage{bm}

\usepackage{graphicx}
\usepackage{epsfig}
\usepackage{enumitem}
\usepackage{url} % not crucial - just used below for the URL
\usepackage{multirow}
\usepackage{epstopdf}
\usepackage{mathtools}

\newcommand{\fullset}{\mathcal{F}}
\newcommand{\informsOR}{1}
\newcommand{\informsMOR}{0}

\graphicspath{{}}

\newtheorem{theorem}{Theorem}
\newtheorem{lemma}{Lemma}

\newtheorem{corollary}{Corollary}
\newtheorem{remark}{Remark}
\providecommand{\keywords}[1]{\textit{\quad Key words }:  #1}

\begin{document}

\title{Best-Subset Selection in Generalized Linear Models: A Fast and Consistent Algorithm via Splicing Technique}

% \author{Yanhang Zhang\thanks{School of Statistics,
% Renmin University of China,
% \texttt{zhangyh98@ruc.edu.cn}}\ ~~~~
% Junxian Zhu \thanks{Saw Swee Hock School of Public Health, National University of Singapore,
% \texttt{junxian@nus.edu.sg}}\~~~~
% Jin Zhu \thanks{Southern China Center for Statistical Science,
% Department of Statistical Science,
% School of Mathematics,
% Sun Yat-Sen University,
% \texttt{zhuj37@mail2.sysu.edu.cn} }~~~~
% Xueqin Wang\thanks{
% Department of Statistics and Finance/International Institute of Finance,
% 	School of Management,
% 	University of Science and Technology of China,
% 	\texttt{wangxq20@ustc.edu.cn} }}

\author[1]{Junxian Zhu\textsuperscript{*}}
\author[2]{Jin Zhu\textsuperscript{*}}
\author[4]{Borui Tang}
\author[2]{Xuanyu Chen}
\author[3]{Hongmei Lin\textsuperscript{$\dagger$}}
\author[4]{Xueqin Wang\textsuperscript{$\dagger$}}

\affil[1]{\footnotesize Saw Swee Hock School of Public Health, National University of Singapore}
\affil[2]{\footnotesize Southern China Center for Statistical Science, Department of Statistical Science, School of Mathematics, Sun Yat-Sen University}
\affil[3]{\footnotesize School of Statistics and Information, Shanghai University of International Business and Economics \\
  \texttt{hmlin@suibe.edu.cn}}
\affil[4]{\footnotesize Department of Statistics and Finance/International Institute of Finance, School of Management, University of Science and Technology of China \\
 	\texttt{wangxq20@ustc.edu.cn} }

\date{}
\maketitle \sloppy

\begin{abstract}
  In high-dimensional generalized linear models, it is crucial to identify a sparse model that adequately accounts for response variation. Although the best subset section has been widely regarded as the Holy Grail of problems of this type, achieving either computational efficiency or statistical guarantees is challenging. In this article, we intend to surmount this obstacle by utilizing a fast algorithm to select the best subset with high certainty. We proposed and illustrated an algorithm for best subset recovery in regularity conditions. Under mild conditions, the computational complexity of our algorithm scales polynomially with sample size and dimension. In addition to demonstrating the statistical properties of our method, extensive numerical experiments reveal that it outperforms existing methods for variable selection and coefficient estimation. The runtime analysis shows that our implementation achieves approximately a fourfold speedup compared to popular variable selection toolkits like \textsf{glmnet} and \textsf{ncvreg}.
\end{abstract}


\begingroup\renewcommand\thefootnote{*}
\footnotetext{Equal contribution}
\begingroup\renewcommand\thefootnote{$\dagger$}
\footnotetext{Corresponding author}

\section{Introduction}
Generalized linear models \citep*[GLMs, ][]{mccullagh1989generalized} can model a family of continuous or discrete responses with a set of predictors.
% It is a broad term encompassing various regression models, including linear regression, logistic regression, and gamma regression.
It generalizes linear regression by assuming responses follow an exponential distribution with a mean determined by a linear combination of predictors.
Specifically, let $\{(\boldsymbol{x}_i, y_i)\}^{n}_{i=1}$ be a dataset with $n$ observations
in which $y_i$ is a response and $\boldsymbol x_i = (x_{i1}, \ldots, x_{ip})^\top$ is
a $p$-dimensional predictor vector.
GLMs assume the density function of $y_i$ given $\boldsymbol x_i$ is:
$f(y_i;\boldsymbol{x}_i^\top \boldsymbol{\beta},\phi) = \exp\{y_i\boldsymbol{x}_i^\top \boldsymbol{\beta} - b(\boldsymbol{x}_i^\top \boldsymbol{\beta}) + c(y_i,\phi)\}$,
where $\boldsymbol \beta \in \mathbb{R}^p$ is a regression coefficient vector,
$b(\cdot)$ and $c(\cdot)$ are some suitably chosen known functions
such that $b'(\boldsymbol{x}_i^\top \boldsymbol{\beta}) = \mathbb{E}[y_i|\boldsymbol {x}_i]$.
% $g(\mu_i)=\theta_i$ is the link function, and $\phi$ is a known scale parameter.
% Here, we assume that there is no intercept effect on $\theta_i$, and we keep this form throughout the paper to simplify presentations.
It is worth noting that $b(\cdot)$ is the critical component in determining the distribution of $y_i$;
% Different distributions and link functions lead to different GLMs.
for instance, let $\theta = \boldsymbol{x}_i^\top \boldsymbol{\beta}$,
taking $b(\theta)=\frac{1}{2}\theta^2$ results in the ordinary linear model
$y_i = \boldsymbol{x}_i^\top \boldsymbol{\beta} + \epsilon_i$, where $\epsilon_i \sim \mathcal{N}(0, 1)$.
Due to their flexibility, GLMs include many frequently-used models like logistic regression, Poisson regression, gamma regression, etc.
Therefore, GLMs adapt to various scientific and engineering disciplines,
including economics, biology, and geography.

Scientific research and data analysis in the modern era frequently collect a large number of predictors for GLMs in order to model responses,
i.e., $p$ is large in comparison to $n$.
It is critical to select a minimally adequate subset of predictors to fit responses accurately,
as this results in a sparse GLM that is accurate and interpretable.
Besides, when $p \approx O(n)$ or $p \gg O(n)$,
minimizing the negative log-likelihood function for $\boldsymbol \beta$,
$$l_n(\boldsymbol \beta)= - \sum_{i=1}^n\{y_i\boldsymbol {x}_i^\top \boldsymbol \beta - b(\boldsymbol {x}_i^\top\boldsymbol \beta) + c(y_i,\phi)\},$$
yields an estimated coefficient with significant fluctuation or even no unique solution.
At this time, pursuing a sparse regression coefficient vector is indispensable.
%Denote the true sparse regression coefficient as $\boldsymbol{\beta}^*$ and its support set as $\mathcal{A}^*$ (also known as the best subset or true active set) with size $s^* \coloneqq |\mathcal{A}^*|$, where $\mathbb{E}[y_i|\boldsymbol {x}_i] = b'(\boldsymbol{x}_i^\top \boldsymbol{\beta}^*)$.
We seek an efficient algorithm to solve the following best-subset selection problem:
\begin{equation}\label{eq:best-subset-glm}
\hat{\boldsymbol \beta} \leftarrow \arg\min\limits_{\boldsymbol \beta} l_n(\boldsymbol \beta), ~\textup{subject to:}~ \|\boldsymbol \beta\|_0 \leq s,
\end{equation}
%such that $\textup{supp}(\hat{\boldsymbol \beta}) = \mathcal{A}^*$.
where $\|\boldsymbol\beta\|_0 = \sum\limits_{j=1}^p \mathrm{I} (\beta_j \neq 0)$ and $\mathrm{I}(\cdot)$ is the indicator function. Since the cardinality constraint on $\boldsymbol \beta$ is non-convex,
finding such an algorithm is not trivial.
% Additionally, $s$ is frequently unknown in practice and must be determined through data.

To this end, the mathematical and statistical communities have spent decades studying
how to select the best subset. One of the most well-known methods is exhaustive enumeration in conjunction with the Akaike/Bayesian information criterion \citep{hocking1967selection, akaike1998information, schwarz1978estimating, anderson2004model}.
The branch-and-bound algorithm improves exhaustive searching by rejecting suboptimal subsets without direct evaluation \citep{lamotte1970computational, narendra1977branch}.
% Motivated by the phenomenal advancements in mixed-integer optimization,
% \citep{bertsimasForumAlgorithmicApproach2016},
% \citet{logistic2017dimitris} and
% \citet{sato2016feature} suggest using
% mixed-integer nonlinear optimization to find the optimal subset in logistic regression with hundreds of variables.
% Very recently, {\color{red}\citet{antoine2021l0learn, bertsimas2021sparse} developed algorithms for
% the best-subset selection under logistic regression models with tens of thousands of variables.}
Unfortunately,
% {\color{red}in the worst cases},
this method continues to impose a sharp increase in computational cost when $p$ grows, as determining the exact best subset is NP-hard, even for linear regression \citep{natarajan1995sparse}.
Empirical evidence is that a widely used software implementing this algorithm \citep{calcagnoGlmultiPackageEasy2010} is limited to datasets with no more than 50 variables \citep{wen2017bess}.

Numerous researchers have attempted to circumvent the NP-hardness.
One important direction is to develop statistical relaxation methods that take advantage of continuous penalties that promote sparsity.
The well-known relaxation methods are the least absolute shrinkage and selection operator \citep*[LASSO,][]{tibshirani1996regression},
elastic net \citep{zouRegularizationVariableSelection2005},
Dantzig selector \citep{emmanuelcandesDantzigSelectorStatistical2007}, smoothly clipped absolute deviation \citep*[SCAD,][]{fan2001variable},
minimax concave penalty \citep*[MCP, ][]{zhang2010nearly}, and truncated $\ell_1$ penalty \citep{xiaotong2012tlp}.
The proposed penalty functions have been extended to GLMs \citep{parkL1regularizationPathAlgorithm2007, jamesGeneralizedDantzigSelector2009, fan2011scad}.
While the majority of relaxation methods can be solved in polynomial time, some of them (for example, the LASSO) lack desirable statistical properties
due to the resulting biased estimations \citep{zhang2008sparsity}.

Another primary direction is the development of algorithms for approximating the best-subset solutions.
A well-known approximation strategy is forward stepwise model selection --- iteratively adding a variable that is highly correlated with the current residuals,
referring to the orthogonal matching pursuit in the machine learning community \citep{mallet1993omp, lozano2011gomp}.
Many researchers believe this procedure is excessively greedy and results in an unsatisfactory solution \citep {weisberg2005applied, friedman2009elements}.
To improve the adaptiveness of forward selection,
\citet{blumensathCompressedSensingNonlinear2013} extend
the iterative hard thresholding algorithm under linear models \citep{blumensathIterativeHardThresholding2009}
to nonlinear objective functions.
\citet{bahmaniGreedySparsityconstrainedOptimization2013}
propose a gradient support pursuit method for best-subset selection under logistic regression.
\citet{wen2017bess} develop a primal-dual active set algorithm for logistic regression that builds on the algorithm developed for linear models \citep{huang2017constructive}.
% {\color{red}\citet{bertsimas2020sparse} consider a dual sub-gradient algorithm motivated by the condition on the equivalent of Problem~\eqref{eq:best-subset-glm} and its Boolean relaxation.}
We refer readers to \citet{zhouGlobalQuadraticConvergence2021} for an excellent review of these methods.
Although $\ell_0$-penalized GLM learning problems are not equivalent to the problem~\eqref{eq:best-subset-glm},
some literature designs algorithms starting with $\ell_0$-penalized GLM learning.
For example, \citet{beckSparsityConstrainedNonlinear2013, antoine2021l0learn} develop coordinate descent algorithms for computing the coordinate-wise minimizer of $\ell_0$-penalized regression.
Unfortunately, the methods above have not been demonstrated to be both statistically recovery-guaranteed and computationally efficient for the best-subset selection under GLMs.
For example, the coordinate-wise minimizer cannot guarantee the recovery of the best subset. At the same time, the primal-dual active set algorithm may sink into periodic iteration even for linear models \citep{foucartHardThresholdingPursuit2011}.
% particularly when the size of the best subset is unknown.

\subsection{Our Proposal and Contributions}

Our primary objective is to contribute a reliable, certifiably efficient, best-subset selection algorithm for GLM. To accomplish this, our algorithm promotes the $s$-cardinality selected subset by excluding ``irrelevant'' variables, including the same amount of ``crucial'' variables, both of which are defined by low-cost computational criteria. The inclusion-and-exclusion iteration continues until the model can sufficiently fit the data. This iterative algorithm avoids the tedious task of enumerating all possible fixed-size subsets. More impressively, our theoretical analysis ensures the correct selection of the best subset after a few iterations. The computational efficiency of our algorithm is also reflected by its effortless convergence. At the same time, the other best subset approximated methods require additional assumptions to ensure convergence to the fixed points of sparsity-pursuit operators. We then design an information criterion to determine the optimal $s$ in a data-driven manner. The information criterion evaluates the solutions with different cardinality and chooses the one that makes a reasonable trade-off between model complexity and accuracy. Theoretical analysis demonstrates that the underlying true subset can be identified in polynomial time with high probability by integrating the information criterion.
% Furthermore, our attention to the delicate computational aspects make our proposed algorithms
% comparable (and at times faster) in speed to the fastest
% proxy algorithms (e.g., glmnet and ncvreg).

This section concludes with the following summary of our contributions:
\begin{itemize}
\item We propose a new fast algorithm for solving the problem of best-subset selection in GLMs based on the following:
(i) an efficient technique for iteratively improving the quality of selected subset and
(ii) an information criterion for choosing the size of the selected subset.
A toolkit \textsf{abess} implementing our algorithm is freely available on CRAN at: \url{https://cran.r-project.org/web/packages/abess}.
\item We theoretically demonstrate that the new algorithm is high-speed.
To put it another way, a few steps of algorithmic iterations reach a stable solution with great certainty.
Additionally, its computational complexity is proportional to the sample size and the number of variables, comparable to the most widely used variable selection method for GLMs---the LASSO.
\item We rigorously establish our proposed algorithm's best-subset-recovery property with a high probability in GLMs, which extends the theoretical guarantees for linear models \citep{zhu2020polynomial}. This extension is not trivial because adapting GLMs to various supervised learning tasks complicates log-likelihood analysis with complex forms.
% Upon this property, we conclude the recovery of the best subset when the sparse level is correctly specified.
\item Our simulation studies provide compelling evidence that our method surpasses state-of-the-art methods in terms of subset selection and parameter estimation across various regression models, including logistic regression, Poisson regression, and multiple-response regression. Furthermore, our method exhibits exceptional performance by achieving a remarkable fourfold speedup compared to the LASSO algorithm. %Our real data analysis reveals that the proposal includes several genes influencing the P53 gene's mutation status.
\end{itemize}

\subsection{Organization and Notations}

The rest of the paper is organized as follows. In Section~\ref{sec:methodology}, we propose a novel technique for improving the selected subset in GLMs and design a new iterative best-subset selection algorithm upon this technique. Section~\ref{sec:theorical-properties} establishes our algorithm's theoretical properties, accompanied by high-level proofs. In Section~\ref{sec:efficient-implementation}, we describe the implementation details for the fast computing of our proposal. Section~\ref{sec:simulation} evaluates the proposed method's empirical performance using artificial datasets.
\if1\informsMOR{
We provide proof of our main results in Section~\ref{sec:proofs}.
}\else{
%A real-world data analysis, as described in Section~\ref{sec:real-data-analysis}, demonstrates the practical benefits of our algorithms.
}\fi
The paper concludes with a few remarks in Section~\ref{sec:conclusion-and-discussion}.
\if0\informsMOR{
Due to space constraints, we relegate the detailed proofs of primary theoretical results and additional numerical experiments to Supplementary Material.
}\fi

Below, we define a few useful notations for the content.
For any vector $\boldsymbol \beta = (\beta_1,\ldots,\beta_p)^\top \in \mathbb{R}^p$,
the $\ell_0$-norm of $\boldsymbol \beta$ is defined as $\|\boldsymbol\beta\|_0 = \sum\limits_{j=1}^p \mathrm{I} (\beta_j \neq 0)$ where $\mathrm{I}(\cdot)$ is the indicator function,
and we define the $\ell_q$-norm of $\boldsymbol \beta$ by $\|\boldsymbol\beta\|_q = ( \sum\limits_{j=1}^p |\beta_j|^q)^{1/q}$, where $q\in[1,\infty)$.
Let $\fullset = \{1, \ldots, p\}$, for any set $\mathcal{A}\subseteq \fullset$,
denote $|\mathcal{A}|$ as its cardinality, $\mathcal{A}^c= \fullset\backslash \mathcal{A}$ as the complement of $\mathcal{A}$ and $\boldsymbol\beta_{\mathcal{A}} =(\beta_j, j\in \mathcal{A}) \in \mathbb{R}^{|\mathcal{A}|}$.
The support set for the vector $\boldsymbol\beta$ is defined as $\mathrm{supp}(\boldsymbol\beta) = \{ j : \beta_j\neq 0\}$.
For a matrix $\mathbf{X} \in \mathbb{R}^{n\times p}$,
define $\mathbf{X}_{\mathcal{A}} = (\mathbf{X}_j, j\in \mathcal{A})\in \mathbb{R}^{n\times |\mathcal{A}|}$,
where $\mathbf{X}_j$ is the $j$-th column of $\mathbf{X}$.
For any vector $\boldsymbol t \in \mathbb{R}^p$ and index set $\mathcal{A}$,
we define $\boldsymbol t|_{\mathcal{A}}$ as a $p$-dimensional vector
whose the $j$-th entry $(\boldsymbol t|_{\mathcal{A}})_j$ is equal to $t_j$ if $j\in \mathcal{A}$ and zero otherwise.
For example,
% $\boldsymbol {\hat \beta}|_{\mathcal{A}}$ is the vector who se $j$th entry is $\hat \beta_j$ if $j\in \mathcal{A}$ and zero otherwise;
$\boldsymbol t|_{\{j\}}$ denotes the vector whose the $j$-th entry is $t_j$ and zero otherwise.
We also simplify $\boldsymbol t|_{\{j\}}$ as $\boldsymbol t|_{j}$ for notational convenience. Denote the true sparse regression coefficient as $\boldsymbol{\beta}^*$ and its support set as $\mathcal{A}^*$ (also known as a best subset or true active set) with size $s^* \coloneqq |\mathcal{A}^*|$, where $\mathbb{E}[y_i|\boldsymbol {x}_i] = b'(\boldsymbol{x}_i^\top \boldsymbol{\beta}^*)$.
\section{Algorithm}\label{sec:methodology}
In this section, we present our fast and consistent best-subset selection algorithm. Firstly, we introduce an algorithm for selecting the best subset when the support size is known. Subsequently, we address the scenario where the knowledge of the support size is unavailable and introduce a novel information criterion to determine the optimal support size.

\subsection{Splicing Method For GLM}
% This section derives an iteration algorithm for solving \eqref{eq:best-subset-glm} given a fixed integer $s$.
Imagine we begin with an arbitrary guess $\mathcal{A} \subseteq \{1, \ldots,p\}$ for the best subset with cardinality $|\mathcal{A}|=s$.
Let $\mathcal{I} = \mathcal{A}^c$, and we shall refer to $\mathcal{A}$ and $\mathcal{I}$ as the active set and inactive set, respectively.
% $$l_{\mathcal{A}} = \min_{\beta_{\mathcal{I}}=0} l_n(\beta).$$
The coefficient estimator under $\mathcal{A}$ is:
\begin{align*}\label{eqn:optimal_beta}
\hat{\bm \beta}=\arg\min_{\boldsymbol \beta_{\mathcal{I}}=0} l_n(\boldsymbol \beta).
\end{align*}
% From the definition of $\hat{\boldsymbol{\beta}}$,
% it can be directly known that
% $\hat{\boldsymbol{\beta}}_j \neq 0$ for $j \in \mathcal{A}$ and $\hat{\boldsymbol{\beta}}_j = 0$ for $j \in \mathcal{I}$.
As follows, define the gradient of $l_n({\boldsymbol \beta})$ at $\hat{\bm \beta}$ as:
\begin{align*}
\hat{\boldsymbol d}
= \left.\frac{\partial l_n( \boldsymbol \beta )}{\partial \boldsymbol \beta}\right|_{\boldsymbol \beta = \hat{\boldsymbol \beta}}.
% = -\sum_{i=1}^{n} (y_i - \mathbb{E}[y_i | \mathbf{x}_i]) \mathbf{x}_i
% = \mathbf{X} \mathbf{w}.
\end{align*}
% where $\mathbf{w} = (y_1 - \hat{\mu}_1, \ldots, y_n - \hat{\mu}_n)^\top$.
% Similarly, we have $\hat{\boldsymbol{d}}\neq 0$ for $j \in \mathcal{I}$ and $\hat{\boldsymbol{d}}_j = 0$ for $j \in \mathcal{A}$.
% And thus, the support sets of $\hat{\boldsymbol{d}}$ and $\hat{\boldsymbol{\beta}}$ are complementary.
Following that, we will introduce two critical concepts for deriving our algorithm.
\begin{itemize}
\item Backward sacrifice: the magnitude of discarding the $j$-th variable in $\mathcal{A}$, i.e.,
\begin{equation}\label{eqn:Delta}
\xi^*_j = l_n\big(\hat{\boldsymbol \beta} |_{\mathcal{A} \setminus \{j\}} \big) - l_n(\hat{\boldsymbol \beta}).
\end{equation}
Intuitively, the $j$-th variable in $\mathcal{A}$ associated with a larger $\xi^*_j$ is
more relevant to the response.
\item Forward sacrifice: the magnitude of adding the $j$-th variable in $\mathcal{I}$ into $\mathcal{A}$, i.e.,
\begin{equation}\label{eqn:delta}
\zeta^*_j = l_n(\hat{\boldsymbol \beta}) - l_n(\hat{\boldsymbol \beta} + \hat{\boldsymbol t} |_{j}),
\end{equation}
where $\hat{\boldsymbol t} = \arg\min\limits_{\boldsymbol t} l_n(\hat{\boldsymbol \beta} + \boldsymbol{t} |_{j})$.
As with the backward sacrifice, a larger $\zeta^*_j$ indicates that the $j$-th variable in $\mathcal{I}$ is more critical for modeling the response.
\end{itemize}
However, it is worth noting that the two sacrifices are incomparable due to their association with different support sets.

The computation of forward sacrifices is time-consuming because an iterative algorithm is needed to minimize $l_n(\hat{\boldsymbol \beta} + {\boldsymbol t} |_{j})$ for all $j \in \mathcal{I}$.
We introduce the following approximations to mitigate this.
%\begin{remark}
According to the definition of $\hat{t}_j$ and the Taylor's expansion, we have
$$\hat{t}_j = -\Big( \left.\frac{\partial^2 l_n( \boldsymbol \beta )}{ (\partial \beta_{j} )^2 }\right|_{\boldsymbol \beta = \bar{\boldsymbol \beta} } \Big)^{-1} \left.\frac{\partial l_n( \boldsymbol \beta )}{\partial \beta_{j}}\right|_{\boldsymbol \beta = \hat{ \boldsymbol\beta} },$$
where $\bar{\boldsymbol \beta} = \hat{\boldsymbol \beta} + (1 - a) \boldsymbol t|_j \; (0 < a < 1)$.
Additionally, using Taylor's expansion and simple algebra, for any $j\in \mathcal{I}$, the forward sacrifice \eqref{eqn:Delta} can be expressed as follows:
\begin{align*}
\zeta^*_j
= - \left.\frac{\partial l_n( \boldsymbol \beta )}{\partial \beta_{j}}\right|_{\boldsymbol \beta = \hat{\boldsymbol \beta} } \hat{t}_j - \frac{1}{2}\left.\frac{\partial^2 l_n( \boldsymbol \beta )}{ (\partial \beta_{j} )^2 }\right|_{\boldsymbol \beta = \bar{\boldsymbol \beta}} (\hat{t}_j)^2
\approx \frac{1}{2}\Big( \left.\frac{\partial^2 l_n( \boldsymbol \beta )}{ (\partial \beta_{j} )^2 }\right|_{\boldsymbol \beta = \hat{\boldsymbol \beta} } \Big)^{-1} ( \hat{\boldsymbol d}_j )^2.
\end{align*}
Similarly, for any $j\in \mathcal{A}$, the backward sacrifice \eqref{eqn:delta} can be approximated by
\begin{align*}
\xi^*_j
= - \left.\frac{\partial l_n( \boldsymbol\beta )}{\partial \beta_{j}}\right|_{\boldsymbol \beta = \hat{\boldsymbol\beta} } \hat \beta_j + \frac{1}{2}\left.\frac{\partial^2 l_n( \boldsymbol \beta )}{ (\partial \beta_{j})^2 }\right|_{\boldsymbol \beta = \bar{\bar{\boldsymbol\beta} }} (\hat\beta_j)^2
\approx \frac{1}{2}\left.\frac{\partial^2 l_n( \boldsymbol \beta )}{ (\partial \beta_{j})^2 }\right|_{\boldsymbol \beta = \hat{\boldsymbol \beta}} (\hat {\boldsymbol \beta}_j)^2,
\end{align*}
where $\bar{\bar{\boldsymbol \beta} }= \hat{\boldsymbol \beta}|_{\mathcal{A} \setminus \{j\} } + a^\prime \hat{\boldsymbol \beta} |_j \; (0 < a^\prime < 1).$
% where $ d_j = \left.\frac{\partial l_n( \boldsymbol \beta )}{\partial \beta_{j}}\right|_{\boldsymbol \beta = \hat{\boldsymbol \beta} }$.
We can see that the sacrifices $\xi^*_j$ and $\zeta^*_j$ can be approximated by the weighted squared values of $\hat{\boldsymbol \beta}_{j}$ and $\hat{\boldsymbol d}_{j}$, respectively.
Our algorithm's central idea is to swap ``irrelevant'' variables in $\mathcal{A}$ for
``important'' variables in $\mathcal{I}$, which may result in a higher-quality solution.
This intuitive idea is referred to as the ``splicing'' technique. Specifically, denote
\begin{equation}\label{eqn:approx_sacrifice}
\begin{split}
\xi_j & \coloneqq \left.\frac{\partial^2 l_n( \boldsymbol \beta )}{ (\partial \beta_{j})^2 }\right|_{\boldsymbol \beta = \hat{\boldsymbol \beta}} (\hat {\boldsymbol \beta}_j)^2 ~(\textup{for}~ j \in \mathcal{A}), \\
\zeta_j & \coloneqq \Big( \left.\frac{\partial^2 l_n( \boldsymbol \beta )}{ (\partial \beta_{j} )^2 }\right|_{\boldsymbol \beta = \hat{\boldsymbol \beta} } \Big)^{-1} ( \hat{\boldsymbol d}_j )^2 ~(\textup{for}~ j \in \mathcal{I}),
\end{split}
\end{equation}
given any splicing size $k\leq s$, and
\begin{equation}\label{eqn:splicing_set}
\begin{split}
\mathcal{S}_{k, 1}= \{j\in \mathcal{A}: \sum_{i \in \mathcal{A}} \mathrm{I}( \xi_j \geq \xi_i) \leq k\},
\;
\mathcal{S}_{k, 2}= \{j\in \mathcal{I}: \sum_{i \in \mathcal{I}} \mathrm{I}( \zeta_j\leq \zeta_i) \leq k\},
\end{split}
\end{equation}
% $\mathcal{S}_{k, 1}$ represents $k$ irrelevant variables in $\mathcal{A}$ and $\mathcal{S}_{k, 2}$ represents $k$ relevant variables in $\mathcal{I}$.
splicing $\mathcal{A}$ and $\mathcal{I}$ by swapping $\mathcal{S}_{k, 1}$ and $\mathcal{S}_{k, 2}$, resulting in a candidate active set $\tilde{\mathcal{A}} = (\mathcal{A}\setminus \mathcal{S}_{k, 1}) \cup \mathcal{S}_{k, 2}$.
Denote $\tilde{\mathcal{I}} = (\tilde{\mathcal{A}} )^c$ and $\tilde{\boldsymbol \beta}=\arg\min\limits_{\boldsymbol \beta_{\tilde{\mathcal{I}}}=0} l_n(\boldsymbol \beta)$.
If $l_n(\tilde{\boldsymbol \beta})$ is significantly smaller than $l_n(\hat{ \boldsymbol \beta})$,
then we believe $\tilde{\mathcal{A}}$ surpasses $\mathcal{A}$,
and we should update the active set: $\mathcal{A} \leftarrow \tilde{\mathcal{A}}$.
The active set can be iteratively updated using the splicing technique until no visible reduction in loss is possible. The above argument is summarized in Algorithm~\ref{alg:fbess}.
\begin{algorithm}[htbp]
\caption{\underline{Be}st-\underline{S}ubset \underline{S}election for \underline{GLM} given support size $s$ (\textbf{BESS-GLM})}
\label{alg:fbess}
\begin{algorithmic}[1]
\REQUIRE A dataset $\{(\boldsymbol{x}_i, y_i)\}^{n}_{i=1}$, an initial active set $\mathcal{A}^0$ with $s$ elements,
the maximum splicing size $k_{\max} (\leq s)$, and a threshold $\tau_s$.
\STATE Initialize: $q \leftarrow -1$, $\mathcal{I}^0 \leftarrow (\mathcal{A}^0)^c$,
$\boldsymbol \beta^0 \leftarrow \arg\min\limits_{\boldsymbol \beta_{\mathcal{I}^0}=0} l_n(\boldsymbol \beta)$, and
$\boldsymbol d^{0} \leftarrow \left.\frac{\partial l_n( \boldsymbol \beta )}{\partial \boldsymbol \beta}\right|_{\boldsymbol \beta = \boldsymbol \beta^0 }$.
\REPEAT
\STATE $q \leftarrow q + 1$, and $L \leftarrow l_n(\boldsymbol \beta^q)$.
\FOR {$k=1, \ldots, k_{\max}$}
\STATE $\xi_j \leftarrow \left.\frac{\partial^2 l_n( \boldsymbol \beta )}{ (\partial \beta_{j})^2 }\right|_{\boldsymbol \beta = {\boldsymbol \beta}^q} ({\boldsymbol \beta}^q_j)^2, j\in \mathcal{A}^q$,
and
$\zeta_j \leftarrow \Big(\left.\frac{\partial^2 l_n( \boldsymbol \beta )}{ (\partial \beta_{j} )^2 }\right|_{\boldsymbol \beta = {\boldsymbol \beta}^q} \Big)^{-1}( {\boldsymbol d}^q_j )^2, j\in \mathcal{I}^q$.
\STATE Update the candidate active set via splicing: $\tilde{\mathcal{A}} \leftarrow (\mathcal{A}^q \backslash \mathcal{S}_{k,1}) \cup \mathcal{S}_{k,2}$, where
\begin{align*}
\mathcal{S}_{k,1} = \{j\in \mathcal{A}^q: \sum_{i \in \mathcal{A}^q} \mathrm{I}( \xi_j \geq \xi_i) \leq k\}, \; \mathcal{S}_{k,2} = \{j\in \mathcal{I}^q: \sum_{i \in \mathcal{I}^q} \mathrm{I}( \zeta_j \leq \zeta_i) \leq k\}.
\end{align*}
\STATE Let $\tilde{\mathcal{I}} \leftarrow (\tilde{\mathcal{A}})^c$,
solve $\tilde{\boldsymbol \beta} \leftarrow \arg\min\limits_{\boldsymbol \beta_{\tilde{\mathcal{I}}}=0} l_n(\boldsymbol \beta)$
and compute
$\tilde{\boldsymbol d} \leftarrow \left.\frac{\partial l_n( \boldsymbol \beta )}{\partial \boldsymbol \beta}\right|_{\boldsymbol \beta = \tilde{\boldsymbol \beta} }.$
\IF {$L - l_n(\tilde{\boldsymbol \beta})>\tau_s$}
\STATE $L \leftarrow l_n(\tilde{\boldsymbol \beta} )$, $({\mathcal{A}^{q+1}} , {\mathcal{I}^{q+1}} , {\boldsymbol \beta}^{q+1}, {\boldsymbol d}^{q+1}) \leftarrow (\tilde{\mathcal{A}} ,\tilde{\mathcal{I}} , \tilde{\boldsymbol\beta}, \tilde{\boldsymbol d })$, and break from the \textbf{for} loop.
\ENDIF
\ENDFOR
\UNTIL{$\mathcal{A}^{q+1} = \mathcal{A}^{q}$}
\ENSURE $(\boldsymbol \beta^{q}, \mathcal{A}^{q}, \boldsymbol{d}^{q})$.
\end{algorithmic}
\end{algorithm}

\begin{remark}
Algorithm~\ref{alg:fbess} necessitates repeated splicing until the active set converges. A good initial guess for the active set $\mathcal{A}_0$ would accelerate the convergence of Algorithm~\ref{alg:fbess}.
The output of sure independent screening for GLMs \citep{jianqingfanSureIndependenceScreening2010} provides an intuitive setting for the initial active set $\mathcal{A}_0$.
To be more precise, we perform predictor-wise regression:
\begin{align*}
% (\hat{\boldsymbol{\beta}}_{j, 0}^M, \hat{\boldsymbol{\beta}}_j^M)
% \leftarrow
% \arg\min_{\beta_0, \beta} - \sum_{i=1}^n\{y_i (\beta_0 + x_{ij}\beta) - b(\beta_0 + x_{ij}\beta) + c(y_i,\phi)\},
% \textup{ for } j=1, \ldots, p.
\hat{{\beta}}_j^M
\leftarrow
\arg\min_{\beta} - \sum_{i=1}^n\{y_i x_{ij}\beta - b(x_{ij}\beta) + c(y_i,\phi)\},
\textup{ for } j=1, \ldots, p.
\end{align*}
The initial set $\mathcal{A}_0$ is then set as $\left\{1 \leq j \leq p_{n}:|\hat{{\beta}}_{j}^{M}| \geq \gamma_{s} \right\}$,
with $\gamma_s$ being the $s$-th largest value among $\{ |\hat{{\beta}}_{1}^{M}|, \ldots, |\hat{{\beta}}_{p}^{M}| \}$.
\end{remark}
\begin{remark}
When $s > s^*$, after the algorithm has covered the true active set, further splicing of irrelevant variables reduces the loss slightly but has no effect on subset selection.
The threshold $\tau_s$ is used to prevent such unnecessary splicing. We choose $\tau_s = 0.01 s\log p\log\log n$ in our implementation based on Assumption~\ref{con:loss-reduce-threshold} in Section~\ref{sec:assumptions}.
\end{remark}
% \begin{remark}
% In our implementation, we consider the standard iterative Newton-Raphson least square (IRLS) method
% and gradient descent to optimize the non-linear loss function under a certain support set.
% IRLS is a second-order optimization method and convergence to the minimum in a few iterations,
% yet the computation of the full Hessian matrix is consuming when $s$ is large.
% Therefore, we employ gradient descent to
% solve the optimization problem, equipped with the diagonal elements in the Hessian matrix.
% \end{remark}
\begin{remark}
The maximum splicing size is specified by $k_{\max}$.
% Our numerical result presented in Figure~\ref{fig:simu_sparse_recovery}
Our numerical results indicate that $k_{\max}$ generally has a negligible effect on identifying the relevant variables.
However, $k_{\max}$ makes a tradeoff between the computational time required by the \textup{\textbf{repeat-until}} loop and the \textup{\textbf{for}} loop.
% A large $k_{\max}$ would consider many possible active sets,
% a part of which may not be better than before but consume computational time.
% A small $k_{\max}$ would improve the quality of active sets slowly and increase the splicing time within \textup{\textbf{repeat-until}} loop.
% {\color{red}(
More precisely, a larger $k_{\max}$ considers more possible active sets in a single \textup{\textbf{for}} loop
but may result in less execution of the \textup{\textbf{repeat-until}} loop.
On the other hand, a smaller $k_{\max}$ consumes less time in a single \textup{\textbf{for}} loop but may result in more times of repetitions.
% )}
Our numerical experience suggests $k_{\max} = 2$ or $5$ is a reasonable choice.
\end{remark}
\begin{remark}
Algorithm~\ref{alg:fbess} naturally terminates since the loss monotonously decreases.
More importantly, it runs extremely fast.
It is clear that the computational complexity of Algorithm~\ref{alg:fbess} is dominated by computing the following:
(i) regression coefficients with fixed support sets;
(ii) the gradient of loss function at the estimated coefficients; and
(iii) the diagonal elements of the Hessian matrix of the loss function.
For (i), the standard GLM textbook's iterative reweighted least squares algorithm can solve it in $O(n s^2)$ loops.
According to (ii), the gradient elements are $\{ \sum\limits_{i=1}^n x_{ij} (y_i - b^\prime(\boldsymbol{x}_i^\top \boldsymbol{\beta})) \}_{j=1}^p$,
% where $\mathbf{w} = , \ldots, y_n - b^\prime(\boldsymbol{x}_n^\top \boldsymbol{\beta}))^\top$,
% In terms of (ii), it is shown to be $\mathbf{X} \mathbf{w}$,
% where $\mathbf{w} = (y_1 - b^\prime(\boldsymbol{x}_1^\top \boldsymbol{\beta}), \ldots, y_n - b^\prime(\boldsymbol{x}_n^\top \boldsymbol{\beta}))^\top$,
and thus can be computed in $O(np)$ loops.
For (iii), the diagonal elements equal to $\{ \sum\limits_{i=1}^n b^{\prime\prime}(\boldsymbol{x}_i^\top \boldsymbol{\beta}) x_{ij}^2 \}_{j=1}^{p}$,
and can be computed using $O(np)$ loops.
When combined with the analysis for the number of iterations within the \textbf{\textup{repeat-until}} loop in Theorem~\ref{thm:num_of_iter},
the computational complexity of Algorithm~\ref{alg:fbess} is roughly controlled by $O(n (p + s^2))$.
On the other hand, searching all possible subsets requires $O(n s^2 p^s)$ loops.
\end{remark}
% {\color{red}(\citep{bertsimas2020sparse} propose a polynomial-time algorithm where, at each iteration, a new support of size k is computed. However, unlike in the present paper, it can be arbitrarily different from the support of the previous iteration. The authors should at least cite and comment on the connection between the two methods. Please include it in the numerical benchmark as well.)}
% {\color{blue}
% \begin{remark}
% We would like to highlight the difference between Algorithm 1 and other algorithms (e.g., \citep{bertsimas2020sparse}, \citep{bertsimas2021sparse}) for solving Problem (1). These algorithms select variables associated with the s smallest of vector $(-\frac{\gamma}{2} \alpha ^{\top} X_j X_j^\top \alpha)_{j=1,\ldots,p} = (-\frac{1}{2\gamma } \beta_j^2 ) _{j=1,\ldots,p}$, where $\alpha$ is the dual variable. These magnitudes are similar to the sacrifices of our method in the active set. In contrast to comparing all the variables simultaneously as these algorithms, our algorithm compares the sacrifices of variables in the active set and inactive set, respectively, and then splices the irrelevant variables in the active set and the relevant variable in the inactive set. It is worth noting that the support of these algorithms can be arbitrarily different from the support of the previous iteration. Our method possesses this property if we set $k_{max}=s$.
% \end{remark}}
%\begin{remark}
We conclude this section by discussing the rationale for leveraging the splicing concept.
Splicing has been demonstrated to be a novel technique for solving the best-subset selection problem under a linear model in polynomial time (in terms of $n$ and $p$) and with a high probability \citep{zhu2020polynomial}.
Empirical evidence indicates that splicing can produce high-quality solutions for best-subset regression problems. Indeed, when dimensionality $p = O(10^5)$ and sample size $n = O(10^3)$, it can effectively solve best-subset selection in half a minute and produce sparse coefficient estimation with desirable predictive power \citep{zhu-abess-arxiv}.
However, the algorithm is not a straightforward generalization because theoretical validation of the splicing concept is significantly more complicated than \citet{zhu2020polynomial} for two reasons.
To begin, because the approximations for the sacrifices \eqref{eqn:Delta} and \eqref{eqn:delta} serve as the splicing criteria, it cannot be deduced directly that splicing results in a higher quality subset selection estimation. The second difficulty level stems from the adaptability and potential complexity of the loss function's form. Our rigorous analysis eliminates both difficulties and establishes statistical properties for Algorithm~\ref{alg:fbess} in Lemma~\ref{lemma:consistency1}.

\subsection{Adaptive Best-Subset Selection}
In this part, we design a data-driven procedure to determine the optimal support size $s$. Model selection methods such as cross-validation and information criteria are widely used techniques. Recently, \citet{fan2013tuning} explored generalized information criterion (GIC) in tuning parameter selection for penalized likelihood methods under GLMs. In particular, we introduce a GIC-type information criterion to recovery support size, which is defined as follows:
\begin{equation}\label{eq:gic}
F(\hat{\boldsymbol \beta}) = l_n( \hat{\boldsymbol \beta} ) + |\text{supp}(\hat{\boldsymbol \beta})| \log(p) \log\log n.
\end{equation}
Intuitively speaking, the model complexity penalty term $|\text{supp}(\hat{\boldsymbol \beta})| \log p \log\log n$ in \eqref{eq:gic} is set to prevent over-fitting, the term $\log\log n$ with a slow diverging rate is used to prevent under-fitting. Combining the Algorithm~\ref{alg:fbess} with GIC, we select the support size that minimizes the $F(\hat{\boldsymbol{\beta}})$. Specifically, we consider a series of candidate support sizes, then we conduct Algorithm~\ref{alg:fbess} for each fixed candidate support size and compute the corresponding GIC via E.q.~\eqref{eq:gic}. The support size that minimizes the GIC is chosen. The procedure above is detailed in Algorithm~\ref{alg:abess}. According to Theorem~\ref{thm:consistency4}, Algorithm~\ref{alg:abess} guarantees consistent best-subset selection with high probability.
\begin{algorithm}[htbp]
\caption{Adaptive Best-Subset Selection for GLM (\textbf{ABESS})}
\label{alg:abess}
\begin{algorithmic}[1]
\REQUIRE A dataset set $\{(\boldsymbol{x}_i, y_i)\}^{n}_{i=1}$ and the maximum support size $s_{\max}$.
\STATE $\mathcal{A}^0 \leftarrow \{ \arg\max\limits_j | (\frac{\partial l_n( \boldsymbol \beta )}{\partial \boldsymbol \beta} |_{\boldsymbol \beta = {\boldsymbol 0} })_j | \}$
\FOR {$s = 1, \ldots, s_{\max}$}
\STATE $(\hat{\boldsymbol \beta}_{s}, \hat{\mathcal{A}}_{s}, \hat{\boldsymbol{d}}_s) \leftarrow \textbf{BESS-GLM}\left( \{(\boldsymbol{x}_i, y_i)\}^{n}_{i=1}, \mathcal{A}^0, k_{\max}, \tau_s \right)$.
% \STATE $\textup{GIC}_s \leftarrow F(\hat{\boldsymbol \beta}_s)$
\STATE $\mathcal{A}^0 \leftarrow \hat{\mathcal{A}}_{s} \cup \{ \arg\max\limits_j|(\hat{\boldsymbol{d}}_s)_j| \}$.
\ENDFOR
\STATE Compute the size of the support set that minimizes GIC: $\hat{s} \leftarrow \arg\min\limits_s F(\hat{\boldsymbol \beta}_s)$.
\ENSURE $(\hat{\boldsymbol \beta}_{\hat{s}}, \hat{\mathcal{A}}_{\hat{s}} )$.
\end{algorithmic}
\end{algorithm}
\begin{remark}
According to Condition \ref{con:maximum-size} in Section \ref{sec:assumptions}, a reasonable choice for $s_{\max}$ is $s_{\max} = [(\frac{{n}}{\log p})^{\frac{1}{4}}]$, where $[a]$ returns the nearest integer of $a$.
\end{remark}
\begin{remark}
In Algorithm~\ref{alg:abess}, Algorithm~\ref{alg:fbess} is run on different support sizes, from a small to a large one. Fortunately, it allows exploiting the latest output of Algorithm~\ref{alg:fbess} to construct a better active set to help Algorithm~\ref{alg:fbess} reach convergence with less splicing iteration. Precisely, in Step 4 of Algorithm~\ref{alg:abess}, we combine the selected best subset returned by Algorithm~\ref{alg:fbess} together with the variable with the most significant forward sacrifice and set it as an initial active set for the following support size. It is the so-called ``warm starts initialization''. Both theory and numerical experiments suggest such initialization is remarkably efficient \citep{barutConditionalSureIndependence2016, friedman2010regularization}.
\end{remark}
% \noindent\rule{\textwidth}{1pt}
% \textbf{Algorithm 2} : \textbf{ABESS} Adaptive Best-Subset Selection \\
% \vspace{-0.5cm}
% \noindent\rule{\textwidth}{0.8pt}
% \vspace{-0.5cm}
% \begin{enumerate}
% \item Input: $\boldsymbol X$, $\boldsymbol y$, $\mathcal{A}^0$, $\mathcal{I}^0$, set $s_{\max} = \frac{n}{\log(p)\log\log n}$ %$\hat{\beta}_0=0$ and $\hat{d}_0 = X' y$.
% \item \textbf{For} $s = 1,2,\ldots,s_{\max}$, \textbf{do}
% $$(\hat{\boldsymbol \beta}_{s}, \hat{\mathcal{A}}_{s})= \textbf{GBESS}(s)~\text{or}~ \textbf{FastGBESS}(s)$$
% \textbf{end for}
% \item Compute the minimum information:
% $$\hat{s} = \arg\min_s \log l_n( \hat{\boldsymbol \beta}_s ) + K(\hat{\boldsymbol \beta}_s) \log(p) \log\log n$$
% Output $(\hat{\boldsymbol \beta}_{\hat{s}}, \hat{\mathcal{A}}_{\hat{s}} )$
% \end{enumerate}
% \noindent\rule{\textwidth}{1pt}

\section{Theoretical Guarantees}\label{sec:theorical-properties}
This section starts by presenting assumptions for the theoretical analysis.
The statistical performance guarantees and convergence analysis of our algorithm are depicted in Sections~\ref{sec:statistical-guarantees} and~\ref{sec:computational-properties}, respectively;
followed by high-level proofs in Sections~\ref{sec:sketch}.
Unless otherwise specified, detailed proofs are relegated to\if0\informsMOR{
Supplementary Materials.
}\else{
Section~\ref{sec:proofs}.
}\fi
Without loss of generality, assume that the design matrix $\mathbf{X} = (\boldsymbol{x}_1,\ldots,\boldsymbol{x}_n)^\top$ has $\sqrt{n}$-normalized columns, i.e., $\mathbf{X}_j^{\top}\mathbf{X}_j=n,\ j=1,2,\ldots,p$.

\subsection{Assumptions}
\label{sec:assumptions}
% To ensure identifiability, we consider the following two concepts. We say $\mathbf{X}$ satisfied the spectrum restricted condition (SRC) with order $s$ and spectrum bound $\{m_{s},M_{s}\}$, i.e.,
% \begin{align*}
% 0<m_s\leq \frac{\|\mathbf{X}_{\mathcal{A}}\mathbf{u}\|_2^2}{n\|\mathbf{u}\|_2^2}\leq M_s <\infty,\forall \mathbf{u}\neq 0, \mathbf{u} \in \mathbb{R}^{|\mathcal{A}|}, ~\mathrm{with} ~ \mathcal{A} \subset \fullset ~\mathrm{and} ~ |\mathcal{A}|\leq s.
% \end{align*}
% The spectrum of the off-diagonal sub-matrices of $\mathbf{X}^\top\mathbf{X}$ can be bounded by the sparse orthogonality constant $\nu_s$ defined as the smallest number such that \label{con:off-diagonal-spectrum}
% $$\nu_s \geq \frac{\|\mathbf{X}_\mathcal{A}^\top\mathbf{X}_{\mathcal{B}}\mathbf{u}\|_2}{n\|\mathbf{u}\|_2}, \forall \mathbf{u}\neq 0, \mathbf{u}\in \mathbb{R}^{|\mathcal{B}|}~\mathrm{with} ~ \mathcal{A}, \mathcal{B} \subset \fullset, |\mathcal{A}|\leq s, |\mathcal{B}|\leq s ~\mathrm{and} ~ \mathcal{A}\cap \mathcal{B}= \varnothing.$$

We present assumptions below to establish the theoretical properties of our algorithm.
\begin{enumerate}[label=(A\arabic*), start=1]
\item There exists a constant $c_0 \in (0, 1]$ such that, the function $b(\theta)$ is three times differentiable with $c_0\leq b''(\theta)\leq (c_0)^{-1}$ and $|b'''(\theta)| \leq (c_0)^{-1}$ in its domain. \label{con:bound-variance}
\end{enumerate}
%\begin{remark}
Assumption \ref{con:bound-variance} avoids infinitely large or small variances of the response \citep{rigollet2012kullback} as well as to ensure the existence of the Fisher information for statistical inference \citep{fan2013tuning}. This assumption is valid for several commonly used GLMs, such as linear regression, logistic regression, and Poisson regression, where the coefficient estimation at any active set $\mathcal{A}$ should be bounded for the latter two.
%\end{remark}
\begin{enumerate}[label=(A\arabic*), start=2]
\item For the model error $W_i=y_i-\mathbb{E}[y_i]$ ($i = 1, \ldots, n$),
$W_1, \ldots, W_n$ follow a sub-Gaussian distribution, i.e., there exist universal positive constants $c_1,c_2$ such that
$\mathbb{P}(|W_i|\geq t)\leq c_1\exp\{-c_2t^2\}$ for any $t>0$. \label{con:subgaussian}
\end{enumerate}
%\begin{remark}
Assumption \ref{con:subgaussian} is usually assumed in GLMs \citep{sahandnnegahbanUnifiedFrameworkHighDimensional2012}.
This assumption holds when the response comes from Gaussian, Bernoulli, or other distributions with bound support (e.g., the gamma distribution).
%\end{remark}

Before presenting the following assumption, we introduce the spectrum-restricted condition (SRC).
A matrix $\mathbf{M}$ enjoying the SRC with order $s$ and spectrum bound $\{m_{s},M_{s}\}$ means:
\begin{align*}
0<m_s\leq \frac{\|\mathbf{M}_{\mathcal{A}}\mathbf{u}\|_2^2}{n\|\mathbf{u}\|_2^2}\leq M_s <\infty,\forall \mathbf{u}\neq 0, \mathbf{u} \in \mathbb{R}^{|\mathcal{A}|}, ~\mathrm{with} ~ \mathcal{A} \subset \fullset ~\mathrm{and} ~ |\mathcal{A}|\leq s.
\end{align*}
\begin{enumerate}[label=(A\arabic*), start=3]
\item
$\mathbf{X}$ fulfills the SRC with order $2s$ and spectrum bound $\{m_{2s},M_{2s}\}$. \label{con:src}
\end{enumerate}
%\begin{remark}
Assumption \ref{con:src} is a widely used identifiability condition in the literature on high-dimensional GLMs \citep{shi2019linear} and was used to investigate the theory of LASSO and MCP \citep{zhang2008sparsity, zhang2010nearly}.
\ref{con:src} implies that the spectrum of the off-diagonal sub-matrices of $\mathbf{X}^\top\mathbf{X}$ is bounded by a constant $\nu_s$ that
is defined as the smallest value such that \label{con:off-diagonal-spectrum}
$$\nu_s \geq \frac{\|\mathbf{X}_\mathcal{A}^\top\mathbf{X}_{\mathcal{A}^\prime}\mathbf{u}\|_2}{n\|\mathbf{u}\|_2}, \forall \mathbf{u}\neq 0, \mathbf{u}\in \mathbb{R}^{|\mathcal{A}^\prime|}~\mathrm{with} ~ \mathcal{A}, \mathcal{A}^\prime \subset \fullset, |\mathcal{A}|\leq s, |\mathcal{A}^\prime|\leq s ~\mathrm{and} ~ \mathcal{A}\cap \mathcal{A}^\prime= \varnothing.$$
% Assumption \ref{con:off-diagonal-spectrum} is the sparse orthogonality constant and
% Assumption \ref{con:src} is the sparse restricted assumption \citep{zhang2008sparsity}.
%\end{remark}
\begin{enumerate}[label=(A\arabic*), start=4]
\item Let
$C_1 \coloneqq \frac{1}{c_0^2}\left(\frac{M_s}{2c_0} + \frac{\nu_s^2}{m_sc_0^3} + \frac{M_sv_s^2}{2m_s^2c_0^5}\right)\left[( 2 + \Delta+ \frac{2\nu_s}{m_sc_0^2})^2 + (2+\Delta)^2\right]\left(\frac{\nu_s}{m_sc_0^2}\right)^2$
and $C_2 \coloneqq \frac{m_sc_0}{2} - \frac{\nu_s^2}{m_sc_0^3} - \frac{M_sv_s^2}{2m_s^2c_0^5}$,
there exists a constant $\Delta > 0$ such that $\gamma_s\coloneqq \frac{C_1}{(1-\Delta)C_2}<1$.
% , $s\leq s_{\max}$. Note that by the monotonicity of $\gamma_s$ on $s$, it suffices to have $\gamma_{s_{\max}}<1$.
\label{con:technical-assumption}
\end{enumerate}
%\begin{remark}
\ref{con:technical-assumption} is a technical assumption. Roughly speaking, we expect $\nu_s$ to be reasonably small for its establishment. Here we derive sufficient conditions in a simpler form for \ref{con:technical-assumption}.
By \ref{con:src} and Lemma~20 in \citet{huang2017constructive}, we have $\nu_s \leq \kappa_s \coloneqq \max \{1-m_{2s}, M_{2s}-1\}$. As a result, it suffices to have $f(c_0,\,\kappa_s) > 0$, where
% and the intuition behind it are given below.
% which is closely related to the the restricted isometry property \citep{candes2005decoding} constant $\delta_{2s}$ for $\mathbf{X}$.
\begin{align*}
f(c_0,\,\kappa_s)
&\coloneqq \frac{(1-\kappa_s)c_0}{2}-\frac{\kappa_s^2}{(1-\kappa_s)c_0^3}-\frac{(1+\kappa_s)\kappa_s^2}{2(1-\kappa_s)^2c_{0}^5} \\
\quad-& \frac{1}{c_0^2}\left(\frac{1+\kappa_s}{2c_0}+\frac{\kappa_s^2}{(1-\kappa_s)c_0^3}+ \frac{(1+\kappa_s)\kappa_s^2}{2(1-\kappa_s)^2c_0^5}\right)\left[\left(2+\frac{2\kappa_s}{(1-\kappa_s)c_0^2}\right)^2+4\right]\frac{\kappa_s^2}{(1-\kappa_s)^2c_0^4}.
\end{align*}
% Due to $f$'s complicated form, we visualize it as a surface in Figure~\ref{fig:sufficient_condition},
% from which simpler sufficient assumptions like $\{c_0\geq 0.9,\ \kappa_s\leq 0.1\}$ or $\{c_0 = 1,\ \kappa_s\leq 0.183\}$ can be derived.
It holds if sufficient assumptions are made, such as $\{c_0\geq 0.9,\ \kappa_s\leq 0.1\}$ or $\{c_0 = 1,\ \kappa_s\leq 0.183\}$. The latter, in particular, corresponds to the case of linear models, and the sufficient assumption $\kappa_s\leq 0.183$ is identical to its counterpart in Remark 2 in \citet{zhu2020polynomial}.

%\end{remark}
% \begin{figure}[htbp]
% \centering
% % \includegraphics[scale=0.1]{figure/sufficient_assumption.png}
% \vspace{-30pt}
% \includegraphics[scale=0.6]{figure/sufficient_condition.pdf}
% \vspace{-30pt}
% \caption{Visualization of $f(c_{0},\kappa_s)$ as a surface, where the $x$-axis is $c_0$, the $y$-axis is $\kappa_s$, and the $z$-axis is $f(c_0, \kappa_s)$.
% The area $\{(c_0, \kappa_s)|f(c_0, \kappa_s)>0\}$ is colored in red, and blue vice versa.
% The two yellow markers are instances when Assumption \ref{con:technical-assumption} is satisfied.}
% \label{fig:sufficient_condition}
% \end{figure}
\begin{enumerate}[label=(A\arabic*), start=5]
\item $\frac{1}{\min\limits_{j\in \mathcal{A}^*} |\beta_j^*|^2} = o \left(\frac{ n}{s\log p\log\log n}\right)$ \label{con:minimal-signal}
\item $\tau_s = \Theta \left({s\log p\log\log n}\right)$ \label{con:loss-reduce-threshold}
\end{enumerate}
Assumption~\ref{con:minimal-signal} sets a minimal magnitude of the regression coefficient and is a commonly employed assumption for variable selection in GLMs \citep{fan2013tuning}. Assumption~\ref{con:loss-reduce-threshold} ensures that the threshold $\tau_s$ can control random errors and prevent unnecessary splicing iterations that reduce loss negligibly. % and enabling necessary splicing iterations (w.r.t upper bound of $\tau_s$).}
%{\color{red}
%\begin{enumerate}[label=(A\arabic*), start=7]
%	\item $k_{\max} = s$ \label{con:k-max}
%\end{enumerate}
%}
\begin{enumerate}[label=(A\arabic*), start=7]
% % \item $|\mathcal{A}^*| = o(\frac{n}{\log p\log\log n})$. \label{con:true-size}
\item $s_{\max} = O\left((\frac{{n}}{\log p})^{\frac{1}{4}}\right)$. \label{con:maximum-size}
\item There exists a constant $c_3$ such that $\max\limits_{1\leq i \leq n}\max\limits_{1\leq j \leq p} |x_{ij}| \leq c_3$. \label{con:x_bound}
\end{enumerate}
\begin{remark}
% Assumption \ref{con:true-size} imposes that the size of the true active set cannot be too large.
% Assumption \ref{con:true-size} can ensure Algorithm~\ref{alg:abess} not to select too small support size.
Assumption \ref{con:maximum-size} controls the growing rate of $s_{\max}$, which ensures that our algorithm avoids selecting an excessively large support size. When establishing model selection consistency on over-fitted models, the Assumption \ref{con:maximum-size} controls the log-likelihood function's third and higher order derivatives. A stronger assumption is presented in Theorem 2 of \citet{fan2013tuning}, where $s_{\max} = O(\min\{ n^{1/6}(\log p)^{-1/3}, n^{1/8}(\log p)^{-3/8}\})$ is assumed for the case of bounded responses. Assumption~\ref{con:x_bound} is a common assumption in high-dimensional GLMs \citep*[see, e.g.,][]{feiEstimationInferenceHigh2021}, which requires that the predictors are uniformly bounded. This assumption is reasonable since preprocessing can be employed to normalize the predictor matrix $\mathbf{X}$. 
% Ignoring the factor $\log \log n$, Assumption \ref{con:minimal-signal} and \ref{con:true-size} require $n$ to scale as $n = \Omega(s_{\max}\log p)$ while
\end{remark}

\subsection{Statistical Guarantees: Support Recovery}\label{sec:statistical-guarantees}
The statistical properties of the support set $\hat{\mathcal{A}}$ returned by Algorithm~\ref{alg:fbess} are presented in the following.
% Theorem~\ref{lemma:consistency1} shows that Algorithm~\ref{alg:fbess} can recover the true support set when the true support size is known.
\begin{lemma}\label{lemma:consistency1}
%{\color{red}Let $\boldsymbol \beta^*$ be the underlying $s^*$-sparse coefficient vector and $\mathcal{A}^*$ be its support set},
When $s^* \leq s$,
under Assumptions \ref{con:bound-variance}-\ref{con:loss-reduce-threshold}, we have
\begin{equation*}
\mathbb{P}(\hat{\mathcal{A}} \supseteq \mathcal{A}^*) \geq 1 - \delta_1 - \delta_2 - \delta_3,
\end{equation*}
where
$\delta_i = O\left(\exp\left\{\log p - K_i\frac{n}{s} \min\limits_{j\in\mathcal{A}^*} |\boldsymbol{\beta}_j^*|^2 \right\}\right)$
for some constant $K_i>0\ (i = 1,\,2,\,3)$.
\end{lemma}
Lemma~\ref{lemma:consistency1} indicates that our algorithm has screening property --- the estimated support set can cover the true support with high probability.
% Theorem~\ref{lemma:consistency1} indicates that Algorithm~\ref{alg:fbess} identifies
% the true support set with high probability.
Furthermore,
Assumption \ref{con:minimal-signal} ensures that $\delta_1,\ \delta_2,\ \delta_3 \rightarrow 0$ as $n \rightarrow \infty$, which leads to the following asymptotic result in Corollary~\ref{corollary:exact-recovery}.
Corollary~\ref{corollary:exact-recovery} guarantees Algorithm~\ref{alg:fbess} covers the true support set with probability 1
when the sample size is sufficiently large,
which can be witnessed by the numerical results displayed in Figure~\ref{fig:rate_binomial}.
\begin{corollary}\label{corollary:exact-recovery}
If assumptions in Lemma~\ref{lemma:consistency1} hold and $s = s^*$, we have
\begin{equation*}
\mathbb{P}(\hat{\mathcal{A}} = \mathcal{A}^*) \to 1, \textup{ as } n \to \infty.
\end{equation*}
\end{corollary}
The establishment of support recovery relies on the idea that,
if some elements of the true active set are omitted,
the splicing procedure improves the quality of the active set and thus
reduce the value of the loss function.
We rigorously prove this statement via proof by contradiction shown in
\if0\informsMOR{
Supplementary Materials.
}\else{
Section~\ref{sec:proofs}.
}\fi
Our idea discriminates from other approaches that design
the sparse-pursuit operators such that its corresponding the fixed point is the true active set \citep{li2017quadratic, yuan2018ghtp, huang2020support}. Their proposal has yet to leverage the information of loss function to decide convergence, and thus, some additional assumptions are to be imposed. Furthermore, the following theorem guarantees Algorithm~\ref{alg:abess} enjoys the consistency of best-subset recovery property, i.e.,
it can recover the true active set with high probability.

\begin{theorem}[Consistency of Best-Subset Recovery]\label{thm:consistency4}
% Assumed
% assumptions in Lemma \ref{lemma:consistency1} hold. If
Suppose
Assumptions \ref{con:bound-variance}-\ref{con:x_bound} hold, then for some positive constant $\alpha > 0$ and sufficient large $n$,
Algorithm~\ref{alg:abess} identifies the true active set (i.e., $\hat{\mathcal{A}}_{\hat{s}} = \mathcal{A}^*$), with probability at least $1 - O(p^{-\alpha})$.
\end{theorem}

\subsection{Computational Properties}\label{sec:computational-properties}
% \begin{theorem}\label{thm:num_of_iter}
% Denote by $(\boldsymbol \beta^q, \mathcal{A}^q)$ the output of the $k$th iteration of Algorithm~\ref{alg:fbess} given support size $s$.
% Suppose that Assumptions \ref{con:bound-variance}-\ref{con:loss-reduce-threshold} hold and $|\mathcal{A}^*|\leq s$.
% Then, with a probability at least $1- \delta_{1}-\delta_{2}-\delta_{3} - \delta_{4}(t)$, we have:
% \begin{enumerate}
% \item[(i)] The total number of iterations needed for structure recovery is upper bounded by
% \begin{align*}
% \mathcal{A}^{*} \subseteq \mathcal{A}^q, \quad \text{if } k > \log_{\frac{1}{\gamma_{s}}}\bigg[\frac{l_n(\boldsymbol 0) + l_n(\boldsymbol\beta^*)}{(1 - \Delta)C_{2}n\min\limits_{j\in \mathcal{A}^*} |\boldsymbol{\beta}_j^*|^2}\bigg];
% \end{align*}
% \item[(ii)] The $\ell_2$-error of parameter estimation is upper bounded by
% \begin{align*}
% \|\boldsymbol \beta^q-\boldsymbol \beta^{*}\|_{2} & \leq \frac{\left[l_n(\boldsymbol 0) + l_n(\boldsymbol\beta^*)\right]\left(\frac{\nu_s}{m_sc_0^2}+1\right)}{(1-\Delta)C_{2}n}\gamma_{s}^q + \frac{t}{m_s c_0}.
% \end{align*}
% \end{enumerate}
% Here, $\delta_{1}$, $\delta_{2}$, $\delta_{3}$ are defined in Theorem~\ref{lemma:consistency1},
% and for any $0<t=O\left(\min\limits_{j\in \mathcal{A}^*} |\boldsymbol{\beta}_j^*|\right)$,
% $\delta_{4}(t) \coloneqq O\left(\exp\left\{\log p - K_4\frac{nt^2}{s}\right\}\right)$.
% % $C_2$ and $\Delta$ are defined in Assumptions \ref{con:technical-assumption}.
% % \begin{align*}
% % \delta_{4}(t): = O\left(\exp\left\{\log p - K_4\frac{nt^2}{s}\right\}\right).
% % \end{align*}
% \end{theorem}
Denote $(\boldsymbol \beta^q, \mathcal{A}^q)$ as the output of the $q$-th iteration of Algorithm~\ref{alg:fbess}. The following theorem characterizes the total number of iterations needed for screening support.

\begin{theorem}\label{thm:num_of_iter}
Suppose Assumptions~\ref{con:bound-variance}-\ref{con:loss-reduce-threshold} hold and $s^* \leq s$.
Then, with probability at least $1- \delta_{1}-\delta_{2}-\delta_{3}$, we have $\mathcal{A}^{*} \subseteq \mathcal{A}^q$ if
\begin{align*}
q > \log_{\frac{1}{\gamma_{s}}}\bigg[\frac{l_n(\boldsymbol 0) + l_n(\boldsymbol\beta^*)}{(1 - \Delta)C_{2}n\min\limits_{j\in \mathcal{A}^*} |\boldsymbol{\beta}_j^*|^2}\bigg],
\end{align*}
where $\delta_{1}$, $\delta_{2}$, $\delta_{3}$ are defined in Lemma~\ref{lemma:consistency1}.
% $C_2$ and $\Delta$ are defined in Assumptions \ref{con:technical-assumption}.
% \begin{align*}
% \delta_{4}(t): = O\left(\exp\left\{\log p - K_4\frac{nt^2}{s}\right\}\right).
% \end{align*}
\end{theorem}
Theorem~\ref{thm:num_of_iter} shows that, within a few
number of iterations, the splicing technique leads to an active set that covers the true active set
(see Proof of Theorem~\ref{thm:num_of_iter} in \if0\informsMOR{Supplementary Material}\else{Section~\ref{sec:proof_num_of_iter}}\fi).
Particularly, a large magnitude of the minimum coefficient lessens
the number of iterations covering the true active set.
% \begin{remark}
% If $s = |\mathcal{A}^*|$, similar to \citet{huang2017constructive}, we can prove Algorithm~\ref{alg:fbess} stops at most $O\Big(\log(|\mathcal{A}^*|R)\Big)$ iteration, where $R = \max\{|\boldsymbol{\beta}_i^*|, i\in \mathcal{A}^*\}/\min\{|\boldsymbol{\beta}_i^*|, i\in \mathcal{A}^*\}$.
% \end{remark}

From Theorem~\ref{thm:num_of_iter}, Algorithm~\ref{alg:fbess} converges with a few iteration.
As such, the proposed Algorithm~\ref{alg:abess} can achieve a polynomial computational time complexity with high probability.
%On the basis of Theorem~\ref{thm:num_of_iter}, the computational complexity of Algorithm~\ref{alg:fbess} is derived below.
%\begin{corollary}\label{thm:complexity}
%	Suppose the assumptions in Theorem~\ref{thm:num_of_iter} hold
%	and Algorithm \ref{alg:fbess} runs with a support size $s$, then
%	\begin{enumerate}
% \item[(i)] If $s<s^*$, its computational complexity is given by
% \begin{align*}
% O\left(\frac{l_n(\boldsymbol 0) + l_n(\boldsymbol\beta^*)}{s\log p\log\log n}(k_{\max}ns^2 + k_{\max}np)\right);
% \end{align*}
% \item[(ii)] otherwise, for $s\geq s^*$, with probability at least $1- \delta_{1}-\delta_{2}-\delta_{3} - O(p^{1-K\log\log n})$ for some constant $K$, Algorithm \ref{alg:fbess} will successfully cover the support with its computational complexity being
% \begin{align*}
% O\left(\left[\log_{\frac{1}{\gamma_{s}}} \frac{l_n(\boldsymbol 0) + l_n(\boldsymbol\beta^*)}{s\log p\log\log n}\right](k_{\max}ns^2 + k_{\max}np)\right).
% \end{align*}
%	\end{enumerate}
%	% \begin{align*}
%	% O\Big(\log_{\frac{1}{\gamma_{s}}} \left[\frac{l_n(\boldsymbol 0) + l_n(\boldsymbol\beta^*)}{s\log p\log\log n}\right]\mathrm{I}(s\geq s^{*}) + \frac{l_n(\boldsymbol 0) + l_n(\boldsymbol\beta^*)}{s\log p\log\log n}\mathrm{I}(s<s^{*})\Big)\times O(k_{\max}ns^2 + k_{\max}np).
%	% \end{align*}
%\end{corollary}
\begin{theorem}\label{thm:complexity}
Suppose Assumptions \ref{con:bound-variance}-\ref{con:x_bound} hold and
Algorithm~\ref{alg:abess} successfully select the true model, i.e., $\hat{\mathcal{A}} = \mathcal{A}^*$
(by Theorem \ref{thm:consistency4}, this event is true with a large probability for sufficiently large sample size $n$),
then the computational complexity of Algorithm~\ref{alg:abess} for a given $s_{\max}\geq s^*$ is
\begin{align*}
O\Big(\big(s^2_{\max}\log_{\frac{1}{\gamma_{s_{\max}}}} \left[\frac{l_n(\boldsymbol 0) + l_n(\boldsymbol\beta^*)}{s^*\log p\log\log n}\right] + \frac{l_n(\boldsymbol 0) + l_n(\boldsymbol\beta^*)}{\log p\log\log n}\big)\big( k_{\max}ns_{\max} + k_{\max}p \big)\Big).
\end{align*}
% Here $\gamma_s$ is defined in Assumptions \ref{con:technical-assumption}.
\end{theorem}
% Note that by Lemma \ref{lemma:consistency1} the uncertainty $1- \delta_{1}-\delta_{2}-\delta_{3}$ is introduced to exclude the event under which Algorithm \ref{alg:fbess} will fail, while an extra term $O(p^{1-K\log\log n})$ is inevitable for controlling the computational complexity.
From Theorem~\ref{thm:complexity}, the proposed Algorithm~\ref{alg:abess} achieves a polynomial computational complexity with high probability---
roughly speaking, the complexity is controlled by
the sample size $n$, the dimensionality $p$, the square of sparsity level $s$,
and logarithmic terms of them.
As such, Algorithm~\ref{alg:fbess} has competitive computational properties for other state-of-the-art variable selection methods such as the LASSO.
The numerical results presented in Figure~\ref{fig:simu_runtime} certify the competitive computational advantage.
% The results is formally established in Theorem~\ref{thm:complexity}.
% \begin{theorem}\label{thm:complexity}
% Suppose assumptions \ref{con:bound-variance}-\ref{con:x_bound} hold and
% Algorithm~\ref{alg:abess} successfully select the true model, i.e., $\hat{\mathcal{A}} = \mathcal{A}^*$
% (by Theorem \ref{thm:consistency4}, this event is true with a large probability for sufficiently large sample size $n$),
% then the computational complexity of Algorithm~\ref{alg:abess} for a given $s_{\max}\geq s^*$ is
% \begin{align*}
% O\Big(\big(s^2_{\max}\log_{\frac{1}{\gamma_{s_{\max}}}} \left[\frac{l_n(\boldsymbol 0) + l_n(\boldsymbol\beta^*)}{s^*\log p\log\log n}\right] + \frac{l_n(\boldsymbol 0) + l_n(\boldsymbol\beta^*)}{\log p\log\log n}\big)\big( k_{\max}ns_{\max} + k_{\max}p \big)\Big).
% \end{align*}
% % Here $\gamma_s$ is defined in Assumptions \ref{con:technical-assumption}.
% \end{theorem}

Finally, we present a theorem demonstrating the gap between estimating the coefficients at the $q$-th iteration and the true coefficients.
\begin{theorem}\label{thm_error_bound}
Under Assumptions~\ref{con:bound-variance}-\ref{con:loss-reduce-threshold},
if $| \mathcal{A}^* | \leq s$, the $\ell_2$-error of parameter estimation is upper bounded by
% \begin{align*}
% \|\boldsymbol \beta^q-\boldsymbol \beta^{*}\|_{2} & \leq \frac{\left[l_n(\boldsymbol 0) + l_n(\boldsymbol\beta^*)\right]\left(\frac{\nu_s}{m_sc_0^2}+1\right)}{(1-\Delta)C_{2}n}\gamma_{s}^q + \frac{\sqrt{\frac{s\log p}{n}}}{m_s c_0},
% \end{align*}
\begin{align*}
\|\boldsymbol \beta^q-\boldsymbol \beta^{*}\|_{2} & \leq \frac{\left[l_n(\boldsymbol 0) + l_n(\boldsymbol\beta^*)\right]\left(\frac{\nu_s}{m_sc_0^2}+1\right)}{(1-\Delta)C_{2}n}\gamma_{s}^q + \frac{t}{m_s c_0},
\end{align*}
with probability at least $1- \delta_{1}-\delta_{2}-\delta_{3} - \delta_{4}(t)$, where
$\delta_{4}(t) \coloneqq O\left(\exp\left\{\log p - K_4\frac{nt^2}{s}\right\}\right)$ for any $t>0$. % =O\left(\min\limits_{j\in \mathcal{A}^*} |\boldsymbol{\beta}_j^*|\right)
\end{theorem}
Theorem~\ref{thm_error_bound} depicts our estimator's $\ell_2$ error bound.
\citet{li2017quadratic} derive an similar parameter estimation error bound in $\ell_2$ norm with order $O(\sqrt{\frac{s\log p}{n}})$.
In fact, for any constant $\delta_4>0$, there exists a constant $K(\delta_4)>0$ (depending only on $\delta_4$) such that if we set $t = K(\delta_4)\sqrt{\frac{n}{s\log p}}$,
then with probability at least $1-\delta_{1}-\delta_{2}-\delta_{3} - \delta_{4}$, our $\ell_2$ error bound attains the same order as that in \citet{li2017quadratic}
when $q\geq \log_{\frac{1}{\gamma_s}} \Big( \sqrt{\frac{n}{s\log p}} \frac{\left[l_n(\boldsymbol 0) + l_n(\boldsymbol\beta^*)\right]\left(\frac{\nu_s}{m_sc_0^2}+1\right)}{(1-\Delta)C_{2}n} \Big)$.

\subsection{A Sketch of the Proofs} \label{sec:sketch}

Here we give a brief high-level description of proofs of the lemma and the four main theorems, pointing out their main ideas and essential ingredients and where the assumptions are used.

To prove Lemma \ref{lemma:consistency1}, we show by contradiction that with a high probability, any support which fails to cover the true support will not be the output of Algorithm \ref{alg:fbess} --- there exists a splicing iteration that can sufficiently decrease the loss function and thus continue the splicing procedure.
Specifically,
let $\mathcal{I}_{1}:=\hat{\mathcal{I}}\cap\mathcal{A}^{*}$ be
the indices of relevant variables that are excluded from the estimated support,
we suppose Algorithm \ref{alg:fbess} fails in support recovery, i.e., $\mathcal{I}_{1}\neq\varnothing$.
On one hand, using Assumptions \ref{con:bound-variance} and \ref{con:src},
the loss under $\hat{\boldsymbol \beta}$ output by Algorithm \ref{alg:fbess} satisfies:
\begin{align}\label{eq:sketch_1}
l_n(\hat{\boldsymbol \beta}) - l_n(\boldsymbol \beta^*) \geq nC_2 \|\boldsymbol \beta^*_{\mathcal{I}_1} \|_2^2 - \varepsilon(\hat{\mathcal{A}}),
\end{align}
where $C_2$ is defined in Assumption~\ref{con:technical-assumption} and
$\varepsilon(\cdot)$ represents error terms converge in probability to $0$ with a convergence rate specified by the sub-Gaussian assumption \ref{con:subgaussian}.
On the other hand, with a well-chosen splicing size $k_0$, we denote the active set obtained after the splicing iteration by $\tilde{\mathcal{A}} = (\hat{\mathcal{A}}\setminus S_{k_0,1}) \cup S_{k_0,2}$.
By the rules of the splicing procedure and again using Assumptions~\ref{con:bound-variance} and \ref{con:src}, we can obtain that
\begin{align}\label{eq:sketch_2}
l_n(\tilde{\boldsymbol \beta}) - l_n(\boldsymbol \beta^*) \leq nC_1 \|\boldsymbol \beta^*_{\mathcal{I}_1} \|_2^2 + \varepsilon(\tilde{\mathcal{A}}),
\end{align}
where $\tilde{\boldsymbol \beta}$ is the minimizer of $l_n(\boldsymbol{\beta})$ under the constraint $\boldsymbol{\beta}_{\tilde{\mathcal{A}}^c} = \mathbf{0}$ and $C_1$ is defined in Assumption~\ref{con:technical-assumption}.
Intuitively, \eqref{eq:sketch_1} and \eqref{eq:sketch_2} indicate that $l_n(\hat{\boldsymbol \beta}) - l_n(\boldsymbol \beta^*)$ and $l_n(\tilde{\boldsymbol \beta}) - l_n(\boldsymbol \beta^*)$ can be roughly approximated by a term proportional to $\|\boldsymbol \beta^*_{\mathcal{I}_1}\|_{2}$ ---
the magnitude of true coefficients that are excluded from the estimated ones.
Combining \eqref{eq:sketch_1}, \eqref{eq:sketch_2} and Assumptions~\ref{con:technical-assumption}-\ref{con:loss-reduce-threshold}, we derive that, with high probability, there will be a contradiction, which completes the proof of Lemma~\ref{lemma:consistency1}.

When $s < s^*$, it can be shown that the difference $l_n(\hat{\boldsymbol \beta}^s) - l_n(\hat{\boldsymbol \beta}^*)$ dominates the penalty term of GIC, i.e. $F(\hat{\boldsymbol \beta}^s) > F(\hat{\boldsymbol \beta}^*)$, where $\hat{\boldsymbol \beta}^s$ is the output of Algorithm 1 given support size $s$. On the other hand, when $s>s^*$, the penalty term of GIC plays the leading role, i.e., $F(\hat{\boldsymbol \beta}^s) > F(\hat{\boldsymbol \beta}^*)$. Combining these two aspects, the GIC attains its minimum at the support size $s^*$.

The proofs of Theorems~\ref{thm:num_of_iter} and \ref{thm_error_bound} rely heavily on intermediate results in the proof of Lemma \ref{lemma:consistency1}. A crucial inequality on which we spend much effort establishing is
\begin{align}\label{eq:sketch_3}
l_n(\boldsymbol \beta^{q})-l_n(\boldsymbol \beta^*) & \leq \gamma_{s}\Big[l_n(\boldsymbol \beta^{q-1})-l_n(\boldsymbol \beta^*)\Big] + R_1,
\end{align}
where $\gamma_s<1$ is a constant defined in Assumption \ref{con:technical-assumption} and $R_1$ is some small remainder term to be controlled. Applying \eqref{eq:sketch_3} recurrently and using
\begin{align}\label{eq:sketch_4}
l_n({\boldsymbol \beta}^{q}) - l_n(\boldsymbol \beta^*) \geq nC_2 \|\boldsymbol \beta^*_{\mathcal{I}_1} \|_2^2 - \varepsilon({\mathcal{A}^{q}})
\end{align}
which is essentially identical to~\eqref{eq:sketch_1}, we can obtain the bound of the number of iterations described in Theorem \ref{thm:num_of_iter}. From Theorem~\ref{thm:num_of_iter}, the loss function decreases drastically as iterations when $s \geq s^*$. When $s<s^*$, the iterations of Algorithm~\ref{alg:abess} can be determined by thresholding to exclude useless splicing. Thus, the computational complexity of Algorithm~\ref{alg:abess} is presented in Theorem~\ref{thm:complexity}.
Finally, using the above inequalities involving $l_n({\boldsymbol \beta})$ for certain $\boldsymbol \beta$ and Assumption~\ref{con:src} that characterizes the curvature of $l_n$, we can naturally derive the upper bound for $\|\boldsymbol \beta^q-\boldsymbol \beta^{*}\|_{2}$ stated in Theorem~\ref{thm_error_bound}.

%As a complementary discussion of Assumption~\ref{con:loss-reduce-threshold},
%we note that only $\tau_s = O \left({s\log p\log\log n}\right)$ is required to establish Lemma~\ref{lemma:consistency1},
%as it avoids the threshold $\tau_s$ being too large to allow necessary splicing iterations.
%One may simply set $\tau_s = 0$ and Algorithm~\ref{alg:fbess} still converges.
%However, to prove its polynomial computational complexity, $\tau_s = \Omega \left({s\log p\log\log n}\right)$ is required to:
%(i) reduce useless iterations when the model is underestimated, and
%(ii) avoid any unnecessary splicing iteration via terminating the algorithm once the true support has been covered,
%as we can see in the proof of Corollary~\ref{thm:complexity}.

\section{Efficient Implementation: Details}\label{sec:efficient-implementation}
We implement the proposed algorithm in a highly efficient \textsf{abess} library
with both Python and R interfaces.
As can be seen from the numerical experiments below,
\textsf{abess} can have competitive or even less running times
compared with other well-known sparse learning software like \textsf{glmnet}.
\textsf{abess} achieves extreme efficiency by leveraging efficient implementation to reduce the time consumption on computing forward and backward sacrifices.
% These include an adaptive grid of tuning parameters, continuation, active set updates, greedy cyclic ordering of
% coordinates, correlation screening, and a careful accounting of floating point operations—some of these
% heuristics (as specified below) appear in prior work
% for deriving highly efficient algorithms for the Lasso (e.g., glmnet).
In the following, we provide a detailed description of the efficient implementation.

\subsection{Simplified Convex Optimization and Early Stopping}\label{sec:approximiated-newton-update}

In Algorithm~\ref{alg:fbess}, we need to solve convex optimization problems:
$\tilde{\boldsymbol \beta} \leftarrow \arg\min\limits_{\beta_{\mathcal{I}} = 0} l_n(\boldsymbol\beta ).$
Directly solving it via a convex optimization solver would consume a large time.
To leverage the sparsity nature of this problem, we can turn to solve a simplified problem:
$$\tilde{\boldsymbol \beta}_{\mathcal{A}} \leftarrow \arg\min\limits_{\boldsymbol{\beta}_{\mathcal{A}}}- \sum_{i=1}^n\{y_i \boldsymbol{\beta}_{\mathcal{A}}^\top (\boldsymbol{x}_i)_{\mathcal{A}} - b(\boldsymbol{\beta}_{\mathcal{A}}^\top (\boldsymbol{x}_i)_{\mathcal{A}}) + c(y_i,\phi)\} $$
and pad zero entries to obtain $\tilde{\boldsymbol \beta}$.
The simplified problem solves the regression coefficient on a much smaller dataset, and thus it is computationally appealing.
Since the simplified problem has no closed-form solution, we must perform iterative algorithms to solve it.
The Newton method is one of the most popular methods to solve this problem.
More precisely, we conduct Newton's updates
% \begin{equation}\label{eqn:primary_fit_update}
% \begin{split}
% {\boldsymbol \beta}_{\mathcal{A}}^{m+1} \leftarrow \boldsymbol \beta_{\mathcal{A}}^m - \Big( \left.\frac{\partial^2 l_n( \boldsymbol \beta )}{ (\partial \boldsymbol \beta_{{\mathcal{A}}} )^2 }\right|_{\boldsymbol \beta = \boldsymbol \beta^m} \Big)^{-1} \Big( \left.\frac{\partial l_n( \boldsymbol \beta )}{ \partial \boldsymbol \beta_{{\mathcal{A}}} }\right|_{\boldsymbol \beta = \boldsymbol \beta^m} \Big),
% \end{split}
% \end{equation}
until $\| {\boldsymbol \beta}_{\mathcal{A}}^{m+1} - {\boldsymbol \beta}_{\mathcal{A}}^{m}\|_2 \leq \epsilon$ or $m > k$,
where $\epsilon$ is convergence tolerance and $k$ is the maximal number of Newton's updates.
Generally, setting $\epsilon = 10^{-6}$ and $k = 80$ returns a desirable coefficient estimation.
% In general, the inverse of the second derivative in \eqref{eqn:primary_fit_update} is computationally expensive, so we use its diagonalized version to approximate it. The update then makes the following changes:
% \begin{equation}\label{eqn:fast_primary_fit_update}
% \begin{split}
% {\boldsymbol \beta}_{\tilde{\mathcal{A}} }^{m+1} \leftarrow \boldsymbol \beta_{\tilde{\mathcal{A}} }^m - \rho D \Big( \left.\frac{\partial l_n( \boldsymbol \beta )}{ \partial \boldsymbol \beta_{\tilde{\mathcal{A}}} }\right|_{\boldsymbol \beta = \boldsymbol \beta^m} \Big),
% \end{split}
% \end{equation}
% where $D = \textup{diag}( (\left.\frac{\partial^2 l_n( \boldsymbol \beta )}{ (\partial \boldsymbol \beta_{\tilde{\mathcal{A}_{1}}} )^2 }\right|_{\boldsymbol \beta = \boldsymbol \beta^m} )^{-1}, \ldots, (\left.\frac{\partial^2 l_n( \boldsymbol \beta )}{ (\partial \boldsymbol \beta_{\tilde{\mathcal{A}}_{|A|}} )^2 }\right|_{\boldsymbol \beta = \boldsymbol \beta^m} )^{-1})$
% and $\rho$ is step size.
% While using the approximation increases iteration time, it avoids a high level of computational complexity when computing the matrix inversion.

Notice that not all of the candidate active sets considered in Algorithm~\ref{alg:fbess} associate with a better solution such that $L - l_{n}(\tilde{\boldsymbol{\beta}}) > \tau_s$,
This inspires us to stop Newton's updates on these active candidates early sets to reduce unnecessary Newton's updates.
More precise, after each Newton's update, we check whether the heuristic criterion
$l_1 - (k - m) \times (l_2 - l_1) > L - \tau_s$ holds,
where $l_1 = l_n({\boldsymbol\beta}_{\mathcal{A}}^{m}), l_2 = l_n({\boldsymbol\beta}_{\mathcal{A}}^{m+1})$.
If so, it implies the remaining $k - m$ times Newton update can potentially lead to a lower loss, and thus,
it has no reason to stop Newton's updates; otherwise, we can stop Newton's updates.
The heuristic criterion is designed based on the fact that the convergence rate of Newton's iterative method is quadratic \citep{nocedalNumericalOptimization1999}.
It considerably improves the efficiency of computing forward sacrifices and
retain the high quality of coefficient estimations.

\subsection{Importance-Priority Splicing}\label{sec:importance-priority-splicing}
Considerable speedup is achieved in a high-dimensional regime by employing an importance-priority splicing procedure,
which is motivated by an active set update for the Lasso \citep{lassoscreening2012tibshirani}.
Specifically, a heavy computational burden comes up when
computing forward sacrifices for $p - s$ variables in the inactive set, especially since $p$ is very large.
To alleviate this burden, after completing forward and backward sacrifices through all the variables, we perform the splicing iteration on a small but essential subset: the union of
the selected variables and $d$ ($\ll p$) variables with the more significant forward sacrifices.
Once the splicing iteration converges, we re-compute sacrifices for all the variables and update the important subset.
If the updated important subset is the same as the previous one,
Algorithm~\ref{alg:fbess} is done. Otherwise, the above procedure is repeated.
From our numerical experience, the procedure mentioned above can save lots of time;
meantime, it keeps Algorithm~\ref{alg:fbess} enjoying the statistical properties
presented in Section~\ref{sec:theorical-properties}.

\section{Simulation Studies}\label{sec:simulation}
In this section, we are mainly interested in the empirical performance of the ABESS algorithm on logistic regression and Poisson regression.
Logistic regression is widely used for classification tasks, and Poisson regression is appropriate when the response is a count.
\if0\informsMOR{In ``Additional Simulation'' of Supplementary Material, we }\else{We }\fi
also consider the performance of ABESS algorithm on multi-response linear regression (a.k.a., multi-task learning).
Before formally analyzing the simulation results,
we illustrate our simulation settings in Section~\ref{subsec:setup}.
% This subsection develops parallel with Section \ref{subsec:logistic}.

%In this section, we study the empirical performance of ABESS for GLM on two generalized linear models,
%logistic regression and gamma regression,
%where logistic regression is widely used for classification and
%gamma regression model is useful for modeling positive continuous response variables.
%Before formally studying logistic regression and gamma regression in Section~\ref{subsec:logistic} and Section~\ref{subsec:gamma}, respectively, we illustrate our simulation setting in Section~\ref{subsec:setup}.

\subsection{Setup}\label{subsec:setup}
To synthesize a dataset, we generate multivariate Gaussian realizations $\boldsymbol{x}_1, \ldots, \boldsymbol{x}_n \overset{i.i.d.}{\sim} \mathcal{MVN}(0,\Sigma)$,
where $\Sigma$ is a $p$-by-$p$ covariance matrix.
%We generate i.i.d error $\epsilon\sim N(0,\sigma^2)$.
%Define the signal to noise ratio (SNR) by $SNR = \frac{\beta^{\top}\Sigma\beta}{\sigma^2}$.
We consider two covariance structures for $\Sigma$: the independent structure ($\Sigma$ is an identity matrix)
and the constant structure ($\Sigma_{ij} = \rho^{I(i\neq j)}$ for some positive constant $\rho$). The value of $\rho$ and $p$ will be specified later.
We set the true regression coefficient $\boldsymbol{\beta}^*$ as a sparse vector with $k$ non-zero entries that have equi-spaced indices in $\{1, \ldots, p\}$.
Finally, given a design matrix $\mathbf{X} = (\boldsymbol{x}_1, \ldots, \boldsymbol{x}_n)^\top$ and $\boldsymbol{\beta}^*$,
we draw response realizations $\{y_i\}_{i=1}^n$ according to the GLMs.

We assess our proposal via the following criteria.
First, to measure the performance of subset selection,
we consider the probabilities of covering true active and inactive sets: $\mathbb{P}(\mathcal{A}^* \subseteq \hat{\mathcal{A}})$ and
$\mathbb{P}(\mathcal{I}^* \subseteq \hat{\mathcal{I}})$ (here, $\mathcal{I}^* = (\mathcal{A}^*)^c$).
We also consider exact support recover probability as $\mathbb{P}(\mathcal{A}^* = \hat{\mathcal{A}})$.
Since the probability is unknown, we empirically examine the proportion of recovery for the active set, inactive set, and exact recovery in 200 replications for instead.
As for parameter estimation performance, we examine relative error (ReErr) on parameter estimations:
$\|\hat{\boldsymbol{\beta}}-\boldsymbol{\beta}^*\|_{2} /\|\boldsymbol{\beta}^*\|_{2}$.
Finally, computational efficiency is directly measured by the runtime.

In addition to our proposed algorithms, we compare classical variable selection methods: LASSO \citep{tibshirani1996regression}, SCAD \citep{fan2001variable}, and MCP \citep{zhang2010nearly}.
%, and a recently proposed coordinate descent (CD) method for $\ell_0$-regularized classification \citep{antoine2021l0learn}.
For all these methods, we apply 10-fold cross-validation (CV) and the GIC to select the tuning parameter, respectively.
% For all these methods, we apply 10-fold cross-validation (CV) to select the tuning parameter.
% ABESS also uses generalized information criterion (GIC) \citep{fan2013tuning} because,
% by combining GIC, ABESS can consistently recover $\mathcal{A}^*$ under linear models \citep{zhu2020polynomial}.
The software for these methods is available at R CRAN (\url{https://cran.r-project.org}).
The software of all methods is summarized in Table~\ref{tab:implementation-details}.
All experiments are carried out on an R environment in a Linux platform with Intel(R) Xeon(R) Gold 6248 CPU @ 2.50GHz. 
\begin{table}[htbp]
\caption{Software for all methods.
The values in the parentheses indicate the version number of R packages.The tuning parameter within the MCP/SCAD penalty is fixed at 3/3.7.}\label{tab:implementation-details}
\centering
\if0\informsMOR{
% \begin{tabular}{ccccccc}
% \toprule
% Method & ABESS & LASSO & SCAD & MCP & CD \\
% \midrule
% Software & \textsf{abess} (0.4.0) & \textsf{glmnet} (4.1-3) & \textsf{ncvreg} (3.13.0) & \textsf{ncvreg} (3.13.0) & \textsf{L0Learn} (2.0.3) \\
% Tuning & sparsity $s$ & $\ell_1$ penalty & $\lambda$ & $\lambda$& $\lambda$ \\
% \bottomrule
% \end{tabular}
\begin{tabular}{cccccc}
    \toprule
    Method & ABESS & LASSO & SCAD & MCP \\
    \midrule
    Software & \textsf{abess} (0.4.0) & \textsf{glmnet} (4.1-3) & \textsf{ncvreg} (3.13.0) & \textsf{ncvreg} (3.13.0) \\
    Tuning & sparsity $s$ & $\ell_1$ penalty & $\lambda$ & $\lambda$ \\
    \bottomrule
    \end{tabular}
}\else{
% \begin{tabular}{ccccccc}
% \hline
% Method & ABESS & LASSO & SCAD & MCP & CD \\
% \hline
% Software & \textsf{abess} (0.4.0) & \textsf{glmnet} (4.1-3) & \textsf{ncvreg} (3.13.0) & \textsf{ncvreg} (3.13.0) & \textsf{L0Learn} (2.0.3) \\
% Tuning & sparsity $s$ & $\ell_1$ penalty & {\color{red}SCAD penalty} & {\color{red}MCP penalty} & $\ell_0$ penalty \\
% \hline
% \end{tabular}
\begin{tabular}{cccccc}
\hline
Method & ABESS & LASSO & SCAD & MCP \\
\hline
Software & \textsf{abess} (0.4.0) & \textsf{glmnet} (4.1-3) & \textsf{ncvreg} (3.13.0) & \textsf{ncvreg} (3.13.0) \\
Tuning & sparsity $s$ & $\ell_1$ penalty & {\color{red}SCAD penalty} & {\color{red}MCP penalty} \\
\hline
\end{tabular}
}\fi
\end{table}
% We implement our proposal in an R package abess \citep{zhu-abess-arxiv}.

\subsection{Logistic Regression}\label{subsec:logistic}

The dimension $p$ is fixed as 500 for the logistic regression model. For the constant correlation case, we set $\rho = 0.4$.
The non-zero coefficients $\boldsymbol{\beta}^*_{\mathcal{A}^*}$ are set to be $(2,2,8,8,8,8,10,10,10,10)^\top$. 
Now we compare methods listed in Table~\ref{tab:implementation-details}.
Figures~\ref{fig:rate_binomial} and \ref{fig:ReErr_binomial} present the results on subset selection and parameter estimation when the sample size increases. Out of clarity, we omit the CV results here and defer these results to the Additional Figures in Supplementary Material.

\begin{figure}[htbp]
\centering
\includegraphics[width=1.0\textwidth]{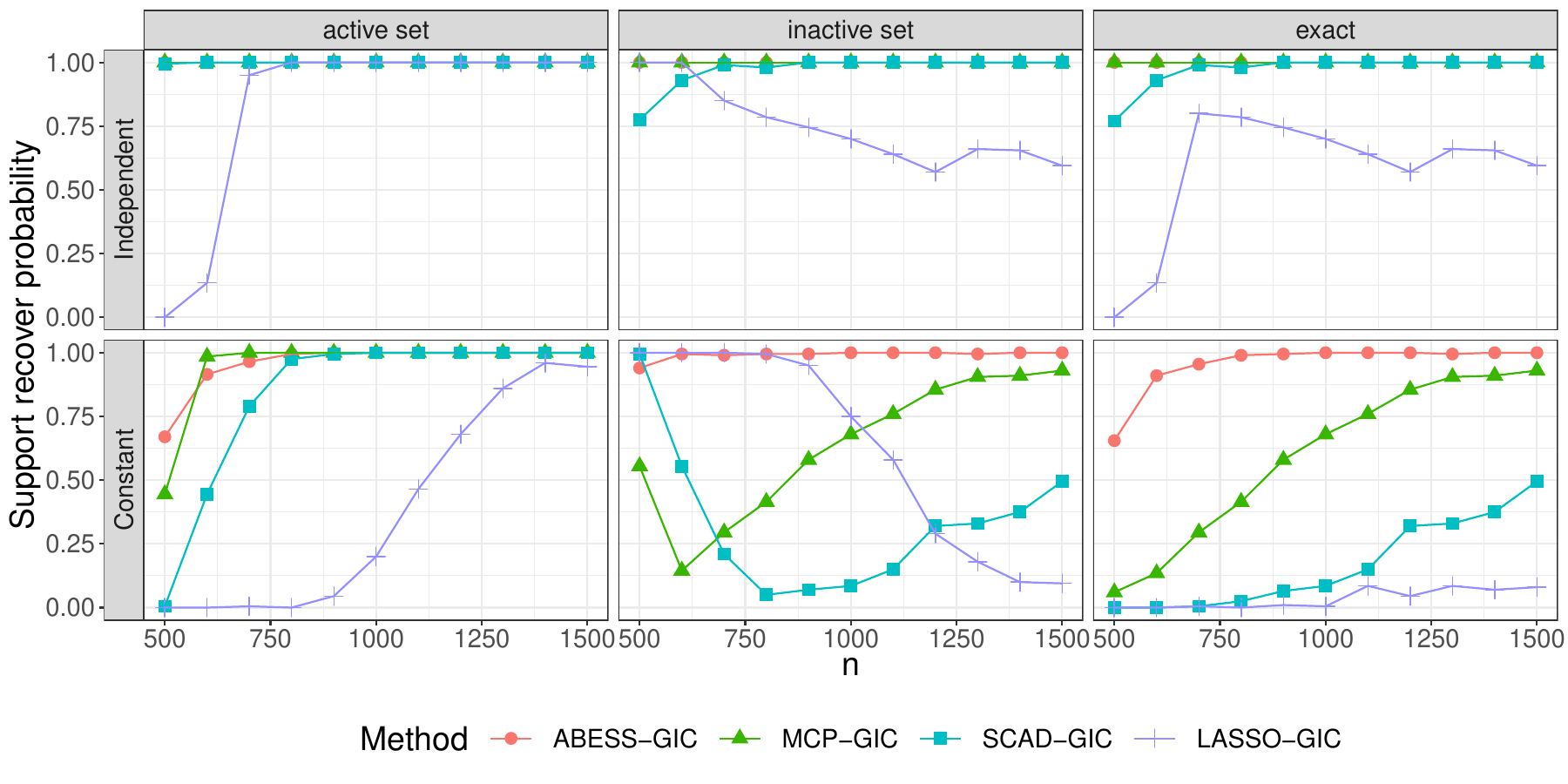}
\if0\informsMOR{
\vspace{-30pt}
}\fi
\caption{Performance on subset selection under logistic regression when covariates have independent correlation structure (Upper) and constant correlation structure (Lower), measured by three kinds probabilities: $\mathbb{P}(\mathcal{A}^* \subseteq \hat{\mathcal{A}})$, $\mathbb{P}(\mathcal{I}^* \subseteq \hat{\mathcal{I}})$, and $\mathbb{P}(\mathcal{A}^* = \hat{\mathcal{A}})$ that are presented in Left, Middle and Right panels, respectively.
}
\label{fig:rate_binomial}
\end{figure}
\begin{figure}[htbp]
\centering
\includegraphics[width=0.8\textwidth]{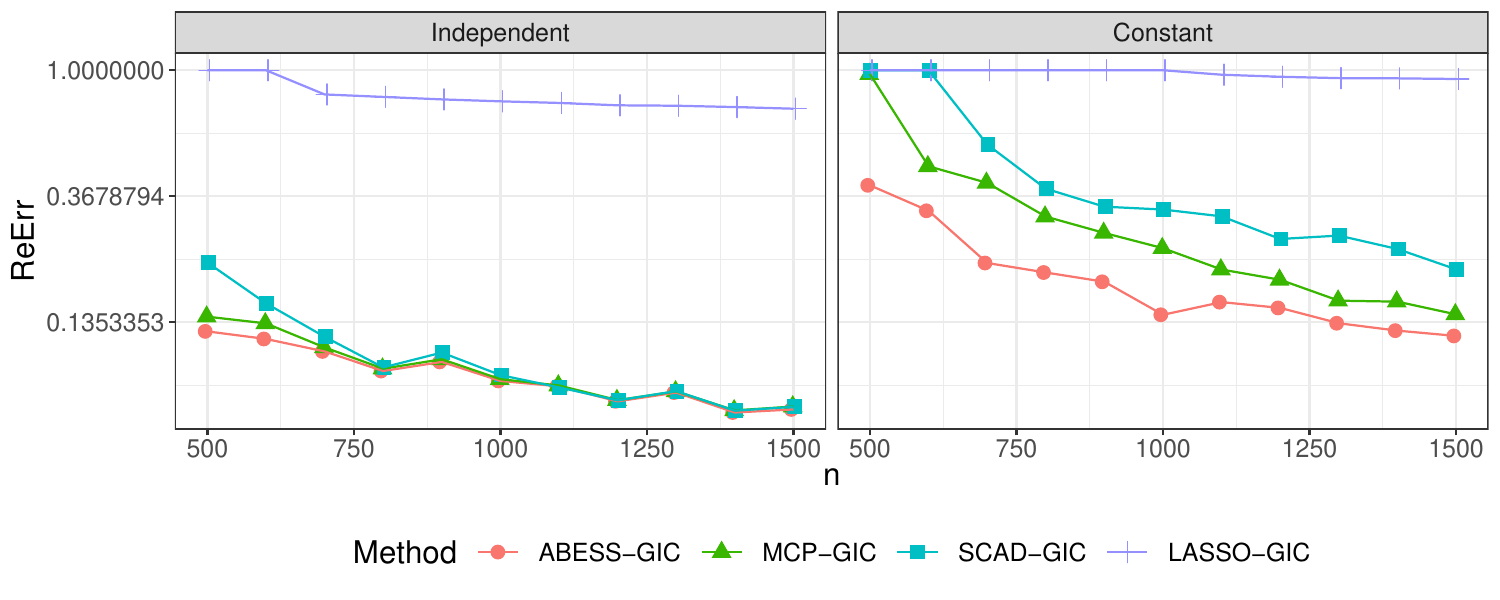}
\if0\informsMOR{
\vspace{-10pt}
}\fi
\caption{Performance on parameter estimation under logistic regression models when covariance matrices have independent correlation structure (Left) and exponential correlation structure (Right). The $y$-axis is the median of ReErr in a log scale.}
\label{fig:ReErr_binomial}
\end{figure}

As depicted in the left panel of Figure~\ref{fig:rate_binomial}, the probability $\mathbb{P}(\mathcal{A}^* \subseteq \hat{\mathcal{A}})$ approaches 1 as the sample size increases, indicating that all methods, except LASSO in the high correlation setting, can provide a no-false-exclusion estimator when the sample size is sufficiently large. However, when considering $\mathbb{P}(\mathcal{I}^* \subseteq \hat{\mathcal{I}})$, as observed in the middle panel of Figure~\ref{fig:rate_binomial}, the LASSO estimator consistently exhibits false inclusions, and the SCAD/MCP estimator shows false inclusions when the covariates are highly correlated. In contrast, only ABESS guarantees that $\mathbb{P}(\mathcal{I}^* \subseteq \hat{\mathcal{I}})$ approaches 1 for large sample sizes. 

Furthermore, as evident from the right panel of Figure~\ref{fig:rate_binomial}, ABESS accurately recovers the true subset under both correlation settings. While SCAD and MCP can also achieve exact support recovery given a sufficient sample size, ABESS demonstrates support recovery consistency with the smallest sample size, particularly when variables are correlated. It is important to note that although our theory imposes restrictions on the correlation among a small subset of variables (see Assumption~\ref{con:technical-assumption}), our algorithm still performs effectively in the constant correlation setting. This setting (i.e., $\rho=0.4$) violates Assumption~\ref{con:technical-assumption} as the correlation between any two variables exceeds 0.183, which is the maximum acceptable pairwise correlation satisfying Assumption~\ref{con:technical-assumption}.

Moving on to Figure~\ref{fig:ReErr_binomial}, it illustrates the superiority of ABESS in parameter estimation. ABESS visibly outperforms other methods in the small sample size regime and maintains highly competitive performance as the sample size increases. This superiority in parameter estimation is not surprising, as ABESS yields an oracle estimator when the support set is correctly identified. Although SCAD and MCP do not provide algorithmic guarantees for finding the local minimum, they exhibit competitive parameter estimation performance due to their asymptotic unbiasedness. Conversely, the LASSO estimator is biased and performs the worst among all the methods.

%\begin{figure}
%	\centering
%	\includegraphics[width=\textwidth]{figure/Performance_binomial.pdf}
%	\caption{Performance comparison under two correlation structures: independent and exponential. (A) Performance for subset selection, measured by support recover probability. (B) Performance for parameter estimation, measured by median ReErr. (C) Average runtime, measured in seconds. L0Learn is omitted since its runtime is far longer than others.}
%	\label{fig:Performance_binomial}
%\end{figure}

\subsection{Poisson Regression}\label{seubsec:poisson}
% As regard to Poisson regression, the response $y_i$ is a integer variable following a Poisson distribution $\mathcal{P}(\lambda_i)$ where $\lambda_i = \exp(\boldsymbol x_i^\top \boldsymbol{\beta})$.
% As a result, the negative log-likelihood is given by
% \begin{equation*}
% l_n(\boldsymbol\beta) = -\sum_{i=1}^{N}\left\{y_{i} \boldsymbol {x}_i^\top \boldsymbol\beta - e^{\boldsymbol {x}_i^\top \boldsymbol \beta} -\log(y_i!)\right\}.
% \end{equation*}
% Empirically, we generate $x_i$ and $\beta$ as described in Section \ref{subsec:setup}.
% Binary response $y_i$ is then drawn from the Bernoulli distribution according to (\ref{eqn:formula_binomial}).
% Let $H_j = \sum\limits_{i=1}^{n} \exp(\boldsymbol {x}_i^\top \hat{\boldsymbol \beta}) x_{ij}^2$ and
% the gradient of $l_n(\boldsymbol{\beta})$ at $\hat{\boldsymbol{\beta}}$ be $\hat{\boldsymbol d} = -\sum\limits_{i=1}^{n}(y_i - \exp(\boldsymbol {x}_i^\top \hat{\boldsymbol \beta})) \boldsymbol {x}_i$,
% \eqref{eqn:approx_sacrifice} can be explicit expressed as:
% $\xi_j = H_j (\hat{\boldsymbol{\beta}}_j)^2$ for $j\in \mathcal{A}$ and
% $\zeta_j = H_j^{-1}( \hat{\boldsymbol d}_j )^2$ for $j\in \mathcal{I}$.
% Given the explicit expression of \eqref{eqn:approx_sacrifice},
% we can conduct Algorithm~\ref{alg:abess} to estimate $\boldsymbol{\beta}$.

For the Poisson regression model, we consider a fixed $p$ value of 500, and set $\rho = 0.2$ for the constant correlation case. The non-zero coefficients $\boldsymbol{\beta}^*_{\mathcal{A}^*}$ are specified as $(1, 1, 1)^\top$. Figures~\ref{fig:rate_poisson_gic}-\ref{fig:ReErr_poisson_gic} present the evaluation of subset selection and parameter estimation quality. Examining Figures~\ref{fig:rate_poisson_gic}, we observe that for ABESS/SCAD/MCP, the probabilities $\mathbb{P}(\mathcal{A}^* \subseteq \hat{\mathcal{A}})$, $\mathbb{P}(\mathcal{I}^* \subseteq \hat{\mathcal{I}})$, and $\mathbb{P}(\mathcal{A}^* = \hat{\mathcal{A}})$ gradually approach 1 as the sample size $n$ increases. In contrary, the LASSO, regardless of the highest inclusion probability for $\mathcal{A}^*$, still has a chance of including ineffective variables, especially when variables are correlated. Comparing ABESS, SCAD, and MCP, it is evident that ABESS achieves the highest exact selection probability, followed by SCAD and MCP. Similar to the results in logistic regression, ABESS achieves exact selection of the effective variables with the smallest sample size under the constant correlation structure.
Regarding the quality of parameter estimation, the ReErr of all methods reasonably decreases as the sample size $n$ increases. Again, ABESS exhibits the least estimation error in terms of the $\ell_2$-norm, which coincides with the results on logistic regression. It is worth noting that our method demonstrates consistency and polynomial complexity under Poisson regression, even though it violates the sub-Gaussian assumption. This is because the current framework of proofs allows for the relaxation of Assumption~\ref{con:subgaussian} to a sub-exponential distribution assumption, enabling the establishment of similar theoretical properties.

\begin{figure}[htbp]
\centering
\includegraphics[width=1.0\textwidth]{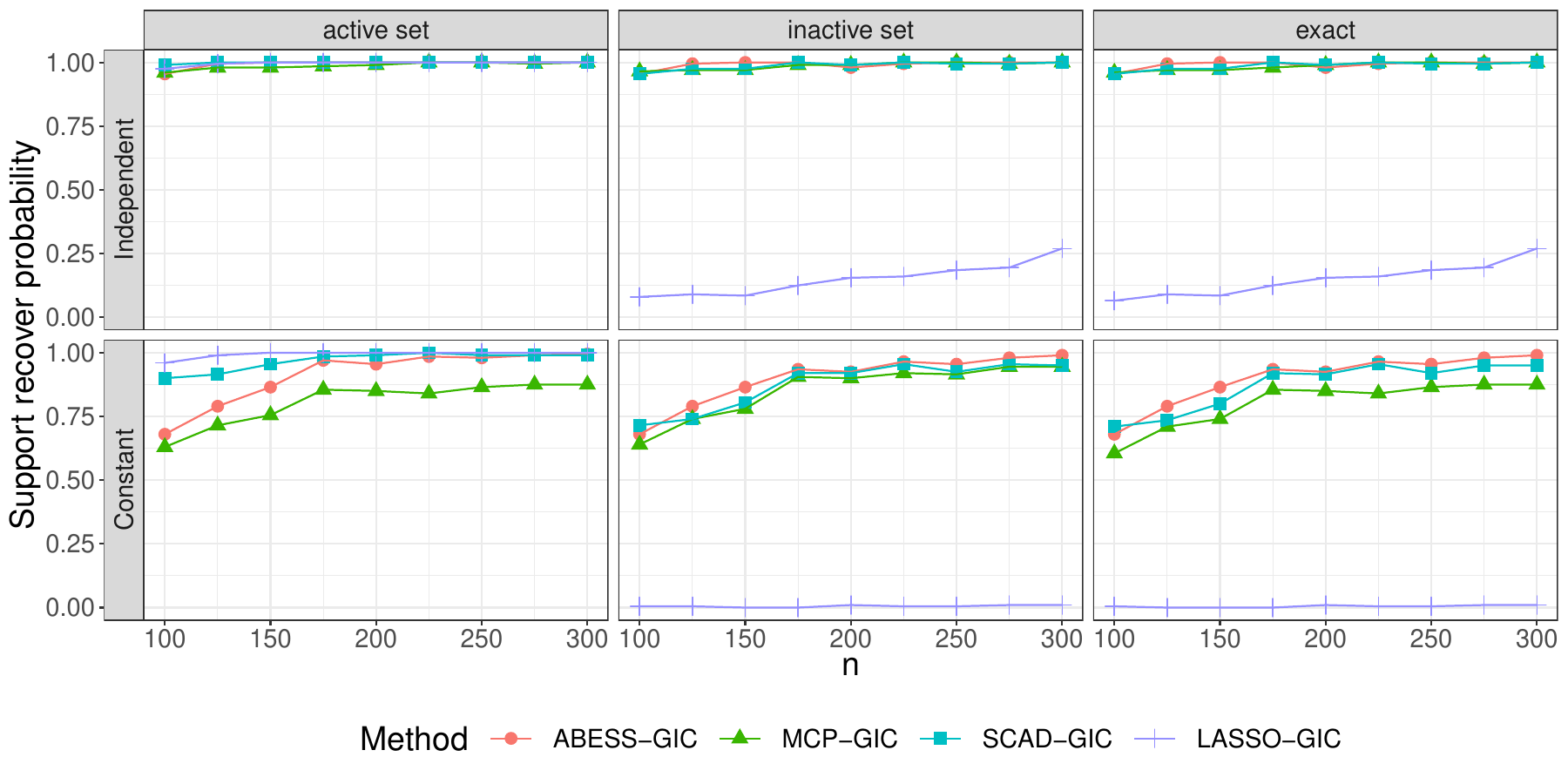}
\if0\informsMOR{
\vspace{-30pt}
}\fi
\caption{Performance on subset selection under Poisson regression when covariates have independent correlation structure (Upper) and constant correlation structure (Lower), measured by three kinds probabilities: $\mathbb{P}(\mathcal{A}^* \subseteq \hat{\mathcal{A}})$, $\mathbb{P}(\mathcal{I}^* \subseteq \hat{\mathcal{I}})$, and $\mathbb{P}(\mathcal{A}^* = \hat{\mathcal{A}})$ that are presented in Left, Middle and Right panels, respectively.}
\label{fig:rate_poisson_gic}
\end{figure}
\begin{figure}[htbp]
\centering
\includegraphics[width=0.8\textwidth]{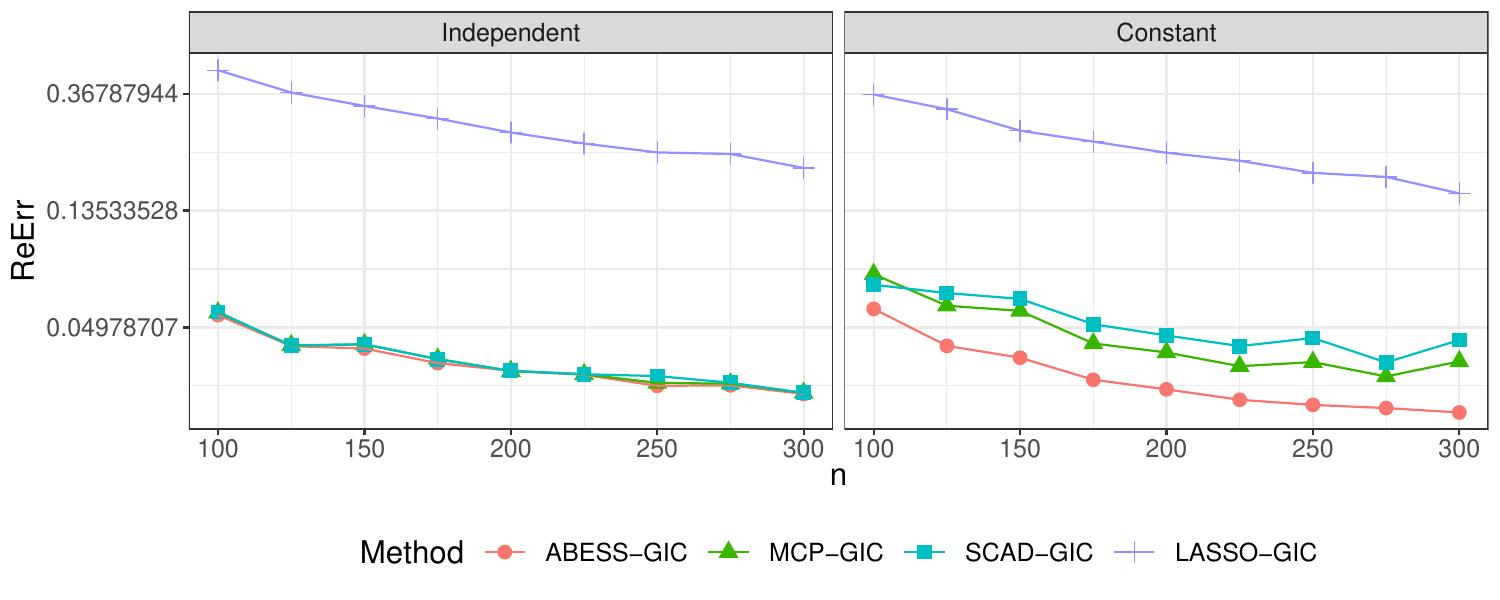}
\if0\informsMOR{
\vspace{-5pt}
}\fi
\caption{Performance on parameter estimation under Poisson regression models when covariance matrices have independent correlation structure (Left) and exponential correlation structure (Right). The $y$-axis is the median of ReErr in a log scale.}
\label{fig:ReErr_poisson_gic}
\end{figure}

\subsection{Computational analysis}

We compare the runtime of different methods in Table~\ref{tab:implementation-details} for the logistic regression and Poisson regression models in Sections~\ref{subsec:logistic} to \ref{seubsec:poisson}. The runtime results are summarized in Figure~\ref{fig:simu_runtime}, indicating that ABESS demonstrates superior computational efficiency compared to state-of-the-art variable selection methods. For instance, when $n = 3000$, ABESS is at least four times faster than its competitors in logistic regression under an independent correlation structure. Furthermore, regardless of logistic regression or Poisson regression, ABESS exhibits similar computational performance, while other competitors run much faster when the pairwise correlation is higher. Lastly, it is important to note that the runtime of ABESS scales polynomially with sample sizes, aligning with the complexity presented in Theorem~\ref{thm:complexity}.
%In contrast, the runtime of other methods grows more rapidly as the sample size increases
%and appears like a quadratic function of the sample size in the independent scenario.
%Increasing iteration numbers for convergence may lead to this result.
%Moreover, ABESS-GIC is faster than ABESS-CV, demonstrating the superiority of the proposed adaptive parameter tuning procedure.
% Finally, according to the computational comparison presented in Figure~\ref{fig runtime_poisson_gic}, the ABESS has the least runtime and is much faster than the MCP and SCAD when variables are independent.

\begin{figure}[htbp]
\centering
\includegraphics[width=0.8\textwidth]{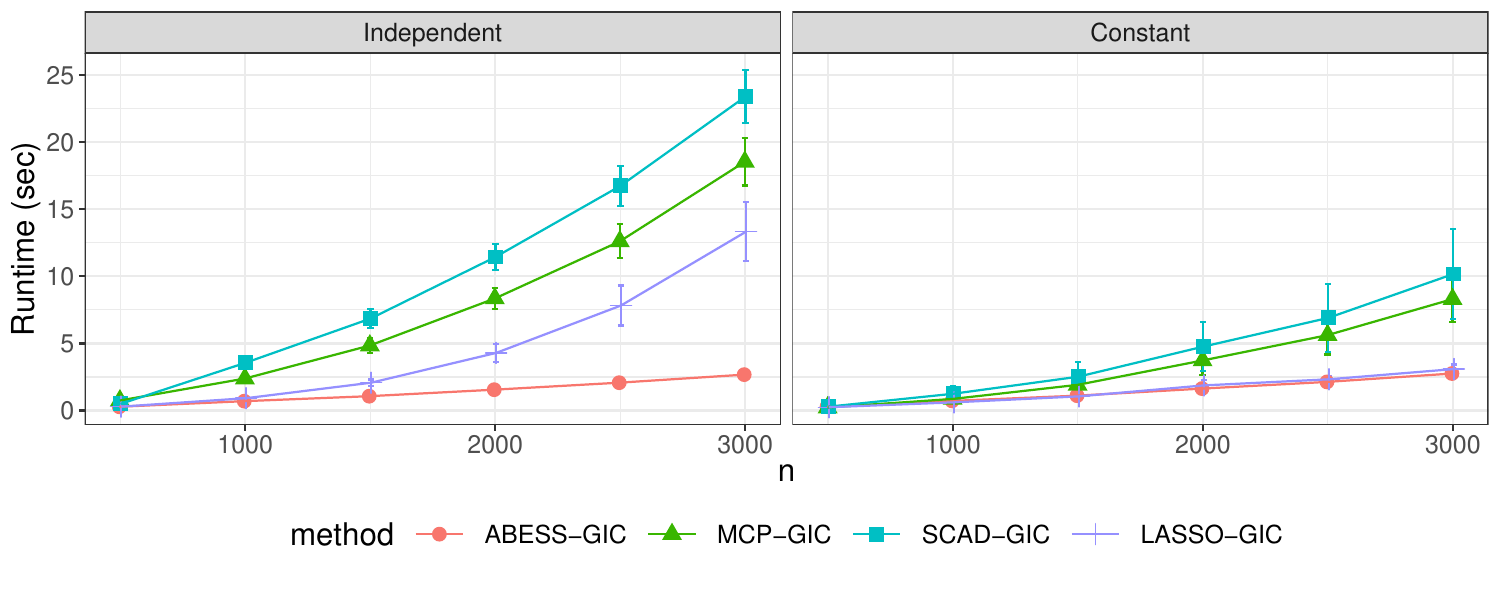}
\includegraphics[width=0.8\textwidth]{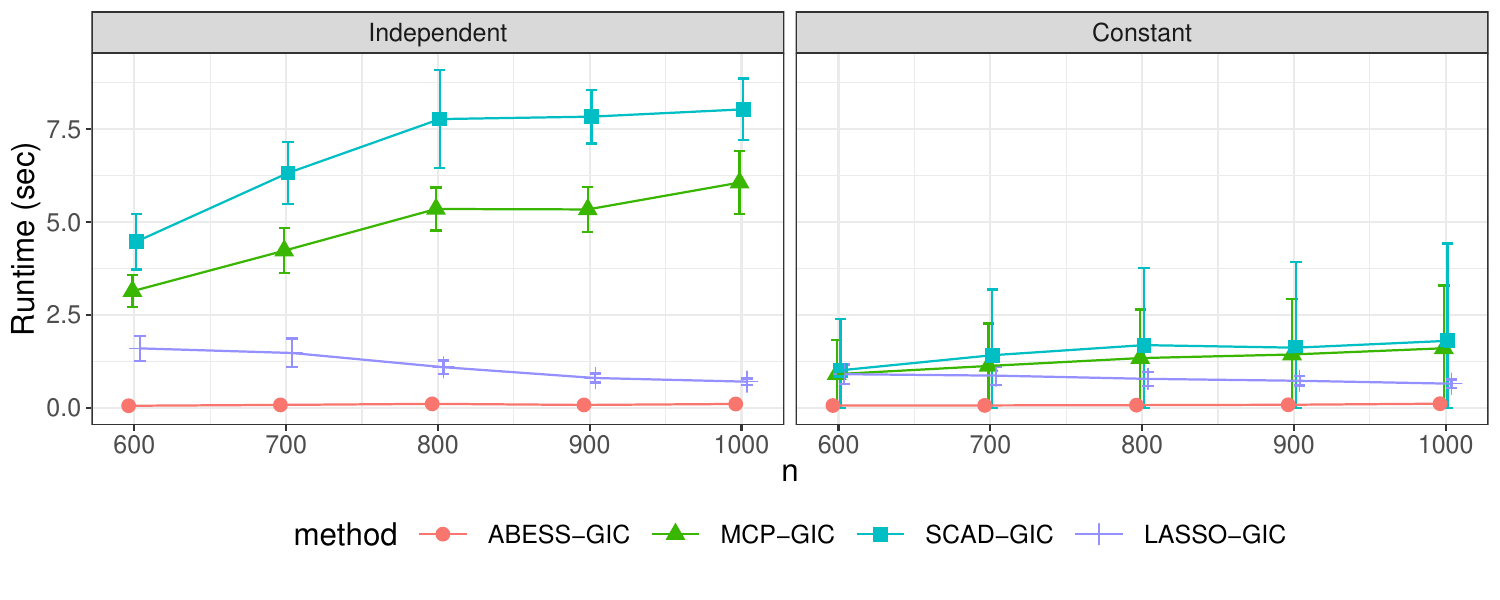}
\if0\informsMOR{
\vspace{-10pt}
}\fi
\caption{Average runtime (measured in seconds) on logistic regression (Upper panel) and Poisson regression (Lower panel). The results on two types of covariances matrix $\Sigma$, the independent correlation structure and constant correlation structure, are presented in the left and right panels, respectively. The error bars represent two times the standard errors.
}
\label{fig:simu_runtime}
\end{figure}

%% For the MOR template, uncomment this line and comment on the code blocks

\if1\informsMOR
{
\input{../appendix_numerical}
}\fi

\section{Conclusion and Discussion}
\label{sec:conclusion-and-discussion}

In this paper, we look for algorithms for the best-subset selection problem in GLM that will be useful to statisticians and practitioners.
We devise novel algorithms to pursue high-quality sparse solutions iteratively, and thus,
circumvent the enumeration of all possible subsets.
More importantly, under mild assumptions, we establish best-subset-screening guarantees and polynomial computational complexity for our algorithms in a high-probability sense.
Our research on logistic regression, Poisson regression, and multi-response regression shows that,
equipped with efficient implementation,
the ABESS algorithm has excellent computational and statistical properties.
The ABESS algorithms are packaged in an openly shared R/Python package abess \citep{zhu-abess-arxiv} that can be used by data scientists worldwide.
% More importantly, the proof technique utilized in this paper can be directly applied to
% more sophisticated scenarios like group variable selection \citep{meierGroupLassoLogistic2008, zhangCertifiablyPolynomialAlgorithm2021}.

% Furthermore, our proposal can be applied to causal inference.
% One of the limitation of our approach is the potential violation of Condition~\ref{con:bound-variance}.
% For example, Condition~\ref{con:bound-variance} might violate if the response come from Poisson distributions.
Several intriguing directions should be investigated further. 
One of our future directions is to apply our algorithmic ideas to a broader range of convex optimization problems
\if0\informsOR
{such as the Cox proportional hazard model and sparse principal component analysis \citep{zouSparsePrincipalComponent2006}.}
\else
{such as the GLMs with latent variables \citep{zhengNonsparseLearningLatent2021}.
}\fi
The application of our proposal to causal inference is another intriguing topic
% \citep{chernozhukovDoubleDebiasedMachine2018, linRegularizationMethodsHighDimensional2015, shortreedOutcomeAdaptiveLasso2017}.
\citep{linRegularizationMethodsHighDimensional2015}.

%%%%%%%%%%%%%%%%%%%%%%%%%%%%%%%%%%%%%
\bibliographystyle{unsrtnat}
\bibliography{ABESS}
%%%%%%%%%%%%%%%%%%%%%%%%%%%%%%%%%%%%%

\end{document}